\documentclass[10pt,twocolumn,letterpaper]{article}

\usepackage{graphicx}
\usepackage{amsmath}
\usepackage{amssymb}
\usepackage{booktabs}
\usepackage{multirow}
\usepackage{adjustbox}
\usepackage{bm}  
\usepackage{subcaption}
\usepackage{caption}
\usepackage{makecell}
\usepackage{wrapfig}
\usepackage[table]{xcolor}  
\newcommand{\etal}{\textit{et al.}}

\newcommand{\graycolor}[1]{\cellcolor{gray!#1}}

\newcommand{\myparagraph}[1]{\vspace{0.1em}\noindent\textbf{#1}}

\usepackage[pagebackref,breaklinks,colorlinks]{hyperref}

\usepackage[capitalize]{cleveref}
\crefname{section}{Sec.}{Secs.}
\Crefname{section}{Section}{Sections}
\Crefname{table}{Table}{Tables}
\crefname{table}{Tab.}{Tabs.}

\begin{document}
	
	\title{Data Pruning in Generative Diffusion Models}
	
	\author{Rania Briq\\
		{\tt\small r.briq@fz-juelich.de}
		\and
		Jiangtao Wang\\
		{\tt\small jian.wang@fz-juelich.de}
		\and
		Stefan Kesselheim\\
		{\tt\small s.kesselheim@fz-juelich.de}
	} 
	\date{Juelich Supercomputing Centre}
	
	\maketitle
	
	%%%%%%%%% ABSTRACT
	\begin{abstract}
		Data pruning is the problem of identifying a core subset that is most beneficial to training and discarding the remainder. While pruning strategies are well studied for discriminative models like those used in classification, little research has gone into their application to generative models. Generative models aim to estimate the underlying distribution of the data, so presumably they should benefit from larger datasets. In this work we aim to shed light on the accuracy of this statement, specifically answer the question of whether data pruning for generative diffusion models (DMs) could have a positive impact. Contrary to intuition, we show that eliminating redundant or noisy data in large datasets is beneficial particularly when done strategically. We experiment with several pruning methods including recent-state-of-art methods, and evaluate over CelebA-HQ and ImageNet datasets. We demonstrate that a simple clustering method outperforms other sophisticated and computationally demanding methods. We further exhibit how we can leverage clustering to balance skewed datasets in an unsupervised manner to allow fair sampling for underrepresented populations in the data distribution, which is a crucial problem in generative models. The source code is available at \href{https://github.com/briqr/diffusion_data_pruning}{GitHub}. 
		
	\end{abstract}
	
	\section{Introduction}
	\label{sec:intro}
	Generative learning has seen tremendous advancement over the past decade, enabling solutions to long-standing challenging problems in image processing such as denoising and upsampling \cite{kawar2022denoising,li2023diffusion,moser2024diffusion}, and found numerous other applications in various domains such as speech, image and video generation\cite{richard2021neural, rombach2021highresolution, ho2022imagen, kondratyuk2312videopoet}, medical image reconstruction and \cite{song2021solving,zhao2023energy,zhang2023generalized,webber2024diffusion} and molecule generation \cite{xu2022geodiff,morehead2024geometry}. This remarkable progress is attributed to sophisticated modeling techniques in deep learning such as Variational autoencoders (VAEs) \cite{kingma2013auto}, Generative Adversarial Networks (GANs)\cite{goodfellow2014generative,gulrajani2017improved,mehralian2018rdcgan}, and more recentaly autoregessive transformer models \cite{kondratyuk2312videopoet,sun2024autoregressive} and diffusion-based models \cite{song2019generative,ho2020denoising, song2020score,dhariwal2021diffusion}. These models rely on the availability of extensive large-scale datasets to achieve high-quality outputs, making them highly demanding in terms of computation power, storage and energy. Consequently, this can often limit their training to only powerful machines with substantial resources therefore rendering  them less widely accessible. Existing research predominantly focuses on model pruning \cite{frantar2023sparsegpt,tao2023structured,santacroce2023matters,sun2023simple}, where the aim is to optimize the generative model's parameters by finding the less critical components and discarding them. These approaches seek to enhance the model efficiency and performance in terms of computational load and memory. However, these efforts concentrate on refining the internal structure of the model rather than addressing data efficiency. In contrast, our attempt to alleviate resource consumption when training generative models shifts the focus towards pruning of the data, through which we aim to identify redundant or harmful samples and retain a core subset that remains sufficiently representative, allowing the model learn most effectively.  Scaling behavior refering to how a given performance metric changes in response to change in the dataset size has been studied in machine translation, for example, in the works \cite{gordon2021data,kaplan2020scaling}, the authors observe that the cross-entropy loss of supervised  machine translation using an autoregressive language model follows a power law, indicating that achieving any improvement on the test error requires exponentially more data. In large language models based on autoregressive transformers, the work of \cite{hoffmann2022training} examines the scaling relationship between a model's size and the amount of training data. It is proposed that both the data and model should be scaled equally to achieve optimality given a compute budget. In contrast, our research is tailored to generative DMs and tries to identify which samples to keep for effective training of these models while keep the model size fixed. 
	Several works have focused on devising pruning methods for discriminative models, including empirical and more theoretically established mthods that provide rigorous theoretical guarantees on the error bound \cite{toneva2018empirical,hwang2020data, coleman2019selection,paul2021deep,mirzasoleiman2020coresets, tan2024data}. Unlike discriminative models which learn decision boundaries to separate the different class categories, generative models aim to model the underlying distribution of the samples. This implies that the benefit obtained by using pruning strategies for learning a decision boundary do not necessarily transfer to a distribution modeling task, as they does not necessarily focus on covering all modes of the distribution, which is essential for training a generative model. This can be explained by the distinct complexity of the loss landscape of generative models in comparison, since they must capture and reproduce intricate patterns within high-dimensional data while using indirect supervision. This higher complexity makes it challenging to understand and infer how different samples contribute during the learning process. For example, while it might be beneficial to prune samples that minimally affect the empirical risk in discriminative models \cite{tan2024data} or samples with a large gradient, it does not necessarily mean that those same samples cannot contribute in learning a distribution. Consequently, when performing data pruning for generative models, distinct considerations must be given to ensure that the reduced dataset still adequately represents the entire data distribution. 
	It may therefore seem that data pruning would harm the performance of a generative model. However, assume that we have lots of noisy or redundant data, if we could selectively train using on the high-quality diverse samples, the model will potentially learn to generate more accurate outputs. For instance, let's imagine a large-scale automotive dataset, such datasets illustare the problem of long-tailed distributions, since much of the data that is collected on the highway is repetitive or uninteresting, but oftentimes we are interested in situations that capture critical scenarios such as a car overtaking another without maintaining a safe distance. Training generative models on such uncurated datasets is therefore not only resource-inefficient, but also detrimental as critical driving scenarios might not be generated in a realistic way. Furthermore, generative models trained on unbalanced datasets pose ethical concerns in society when certain groups are underrepresented, further perpetuating existing inequities \cite{king2022harmful,teo2023fair,teo2024measuring,hao2023safety,cho2023dall,lee2024holistic,masrourisaadat2024analyzing,zhang2018unreasonable,farina2024ethical}. Data pruning  could offer a solution by allowing to control the distribution, guaranteeing balanced distributions so that under-represented populations will be generated equally reliably. In this work, we hope to gain insight into what makes a sample redundant, how to distinguish such a sample in the distribution, what pruning threshold will break the model and we can balance skewed distributions. We will see that diffusion models have a high tolerance threshold for pruning. In fact, on ImageNet, we find that we can prune as much as 90\% of the data without any decline in performance. Furthermore, when using clustering in the embedding space of a large visual model, we achieve better performance than the unpruned variant. This also implies that even in case where we have little data, we can still train a DM and obtain reasonable results.
	
	\section{Related Work}
	\label{sec:related_work}
	Sorting samples by their importance allows to train a discriminative model with a reduced set size while maintaining or minimally affecting their performance, thus saving compute and storage. There exists a body of work that focuses on this problem\cite{har2005smaller, toneva2018empirical,hwang2020data, coleman2019selection,paul2021deep,mirzasoleiman2020coresets, tan2024data,abbas2024effective,xia2022moderate,killamsetty2021grad,kaushal2021prism,zheng2022coverage,yang2022dataset,feldman2020neural,yang2024mind,he2024large}. For example, Toneva \etal \cite{toneva2018empirical} assesses a sample's difficulty by tracking the frequency at which the model flips a correct prediction to an incorrect one. Their empirical findings suggested that samples that are more often forgotten have the most impact on the model. Paul \etal
	\cite{paul2021deep} propose to score samples importance based on the error L2-Norm (EL2N) and the gradient norm (GraNd) early in the training. Feldman \etal \cite{feldman2020neural} score a sample based on the model's ability to memorize and predict its represented subpopulation. Tan \etal \cite{tan2024data} tries to resolve the drawbacks of existing methods such as noisy samples erroneously obtaining high scores despite being harmful. Instead, they propose Moving-one-Sample-out (MoSo) to score samples based on importance rather than difficulty, by estimating the effect of removing a sample on the empirical risk, or alternatively, scoring high a sample whose gradient matches the average gradient of the data. Works such as \cite{sorscher2022beyond,abbas2024effective,xia2022moderate} cluster samples based on their embeddings in a feature space extracted by a strong visual model, and score samples by defining a difficulty metric with respect to the clusters such as distance to the center or region density. He \etal \cite{he2024large} use prediction uncertainty as an indicator for pruning, which measures predictions variation within a sliding window. Yang \etal \cite{yang2024mind} prune samples based on their proximity to the decision boundary, observing that the boundary is most impacted by adjacent samples, as such, selecting those samples ensures that the decision boundary can be reconstructed using a reduced dataset.
	%------------------------------------------------------------------------
	\section{Data pruning for Diffusion models}
	The task of generative modeling primarily centers on developing a sampler for an unknown data distribution $p^*(x)$, using  independently and identically distributed samples drawn from this distribution. Such a sampler should generate novel samples that follow the same underlying distribution. Within this framework, we introduce the concept of data pruning for generative models, in which given a dataset S that contains $n$ data samples used to train a diffusion generative model, find an optimal training subset $S'\subset S$, that preserves or potentially improves its performance. We experiment with various pruning methods with compehensive evaluation of their effect on the diffusion model.
	\subsection{Diffusion models}
	Diffusion models  are a class of generative models that transform simple, noisy data distributions that are easy to sample from into a more complex target distribution $p^*$. By iteratively applying learned transformations, the model reduces the various noise levels added in the forward process until ultimately converging to a real image. The objective during this reverse process is to predict the additive noise through score matching, which guides the mapping to a clean sample. 
	In the following, we expound on the different pruning strategies that we implemented to train a DM model.
	\subsection{Data Pruning methods}
	Below we explain in detail the pruning methods that we applied to select a subset of the training data to train a generative DM. For each method, we also apply its inverse, that is, if for a certain method we select samples with the highest score based on this method's criteria, we base our selection criteria on inverted score and rerun the experiment. This helps to gain more insight into the method's efficacy, for example, if its inverse achieves the opposite effect and degenerate the quality this would indicate that the selection criteria is not effective. When pruning 50\% based on a method and pruning 50\% based on its inverse, this would create two disjoint subsets for training the model in two separate runs. Furthermore, to check whether something in the middle would yield a more balanced distribution, we takes samples in the middle of these two extremes. These experiments comprise our ablative studies in evaluating each pruning method's efficacy. To learn how the model behaves as we vary pruning ration, we run each of these pruning methods using several pruning ratios. 
	
	\myparagraph{Random pruning}. Samples are randomly pruned without being assessed. Before training is started, we sample a random subset and use it to train the model throughout the process. This serves as a baseline against which pruning strategies that require computation are evaluated. If a method performs on par with random pruning, it indicates that the strategy might not be as effective in identifying and retaining the most informative samples for training. 
	
	\myparagraph{Loss Monotonicity}. This pruning method is inspired by \cite{toneva2018empirical}, where the concept of a sample hardness  is based on how many times the network flips its correct classification of the sample during training. This measure can be indicative of the model's uncertainty and difficulty in learning a specific sample. To make this method suitable for DMs, we adapt it based on the monotoncity of the DM objective which involves minimizing the mean square error (MSE) between the additive noise and predicted noise. As the model learns, it is expected that the general behaviour of this objective is monotonically decreasing. This method requires  a short pretraining phase consisting of a small number of epochs to gather data on the loss behavior. For evaluation fairness of a sample across the different epochs in this phase, we fix the time step $t$ in the diffusion process to a small value, so it is representative of the true distribution rather than the perturbed one. Specifically, for each sample during training, we keep track of the frequency at which the loss value increases over its previous value from the earlier epoch for a predeterimed timestep. Samples exhibiting the highest number of  increases are selected. The inverse of this strategy selects samples exhibiting the fewest number of increases . 
	
	\myparagraph{GraNd.} Analogous to \cite{paul2021deep}, in which samples difficulty is  based on the  gradient norm. Specifically, the method calculates the gradient of the objective with respect to the model parameters for each sample. In a classification model this corresponds to the entopy loss, where samples with a larger gradient contribute more to the model's parameters during training. In our context, the score function $\nabla_x \log p_{\theta}(x)$, which is estimated by the gradient with respect to the network parameters $\theta$, describes how the likelihood changes with respect to a sample $x$. Samples in regions with higher density will typically have larger gradients, so retaining these samples should be beneficial for estimating the probability density. This evaluation requires a pretraining phase as well, where in the final epoch we compute the gradient. We also fix the time step $t$. We apply the inverse criteria which indicates samples in less dense regions are selected.
	
	\myparagraph{EL2N}.  In \cite{paul2021deep}, sample importance is based on the $\ell_2$ loss magnitude. The measure scores more high samples it finds challenging as the loss value is indicative how well the model is performing on each sample. In a DM context, this has the opposite connotation, as higher loss indicates lower probability density, synonymous to GraNd-based pruning. It might therefore be more beneficial to score samples inversely. This also requires a short pretraining phase where in the final epoch we record the loss for each sample and select those with the highest/lowest loss value. 
	
	\myparagraph{Moving-one-Sample-out (MoSo)} . A recent pruning method that was proposed by Tan \etal \cite{tan2024data}. In this approach, a sample's importance is determined by its impact on the optimal empirical risk. This impact is measured by the change  a sample removal from the training set incurs on the empirical risk, which requires retraining a model without the sample in question. Evidently this is prohibitively expensive, so the authors propose to evaluate the change using first-order approximation, in which they evaluate the extent to which a sample's gradient agrees with those of other samples by measuring its similarity with the average gradient. This method is more robust to outliers, since samples with high scoring according to \textit{GraNd} could merely be indicative of the presence of noise rather than valuable information. In a DM context, like for GraNd, it asserts samples are constantly in higher density regions. This method also requires a pretraining phase in which several surrogate networks are trained on multiple subsets of the data and are later used to approximate the similarity between a sample's gradient with the average gradient using a dot product. This method is the most computationally expensive amongst the pruning methods that we implemented, so we reserve it only to the smaller-scale dataset in our experiments.
	
	\myparagraph{Clustering}. We use a pretrained powerful backbone such as  CLIP \cite{radford2021learning} and DINO \cite{oquab2023dinov2} to extract the samples features. These models yield embeddings containing rich semantic information about the samples, which we use to cluster the samples, ordering them in groups that share similar semantic characteristics. For each cluster, we score its population based on their distance from the cluster center, by taking either the samples located closest to its center, the furthest, or those in-between.  Samples that are located near the center hold typical features that are represetative of the cluster's population distribution. When aggregating all clusters' nearest samples into a training subset, we effectively have representative features across the spectrum, helping retrain the core characterstics of the distribution. When we select samples located furthest away from the center, we cover more dificult and scarce samples, which exposes the model to unique samples. The number of samples selected from each cluster here is proportional to its size, which means that if the distribution of the original data is imbalanced and certain classes are underrepresentated, the pruned dataset will be equally imbalanced. However, leveraging these clusters, we can easily balance skewed datasets by simply taking an equal number of samples from each cluster. We run this experiment with a single pruning ratio determined by the size of the smallest cluster. We refer to these experiments as Cluster$_{backbone}$, Cluster$_{backbone}^{mid}$ and Cluster$_{backbone}^{-1}$ when taking the closest, middle and furthest samples correspondingly, where backbone refers to either CLIP (C) or Dino (D).
	
	\section{Experiments}
	\myparagraph{Datasets.}
	We train and evaluate our diffusion  model using two datasets of two different scales in order to contrast the effect of pruning at different dataset scales. 
	
	\myparagraph{CelebA-HQ} consists of high-resolution images of diverse human faces of celebrities and contains $28k$ training images  and $\approx 2k$ validation images. When clustering, we set $k=24$ by analyzing the inertia values.
	
	\myparagraph{ImageNet1K} is a comprehensive label-annotated  visual dataset spanning 1000 distinct object categories. It contains $1.2M$ training, $50k$ validation and $100k$ test images correspondingly. The raw images combined consume 133 GB of storage. When clustering, we select the number of clusters to be 1000, corresponding to same number of object categories.
	
	\myparagraph{Evaluation metrics.} Evaluating generative models remains challenging, as there does not exist a single metric that can comprehensively assess the overall performance of generative models. We therefore employ a diverse set of evaluation metrics to throughly examine the impact pruning has on the model across different pruning ratios. The standard  metric is Fréchet Inception Distance \textbf{(FID)}  which measures the Wasserstein distance between the distributions of generated and real images. We also employ the \textbf{Inception} score which evaluates fidelity and class diversity, \textbf{F-score} which is a harmonic average of precision and recall corresponding to fidelity and diversity\cite{sajjadi2018assessing,kynkaanniemi2019improved},  and \textbf{Vendi} score \cite{friedman2022vendi} which is an additional measure to evaluate diversity within a dataset without requiring a reference dataset. It computes the exponential of the Shannon entropy of the eigenvalues of the samples pairwise similarity matrix. Additionally, we attempt to detect memorization where the model is prone to replicating the training samples by employing the simple test proposed in \cite{meehan2020non}, specifically, we extract the generated samples' features and measure the average distance of each sample to its nearest match in the training set.

	\myparagraph{Training}
	We implement our work on top of DiT framework \cite{Peebles2022DiT}, which is a transformer-based architecture using their github repository \footnote{https://github.com/facebookresearch/DiT} . Instead of reversing the diffusion process using their scored-based diffusion model, we use flow matching and sample $x_t$ using a probability flow ordinary differential equation (ODE)\cite{lipman2022flow,ma2024sit} \footnote{https://github.com/willisma/SiT}, which allows to reverse the diffusion process using fewer sampling steps.  We also train our own Vector Quantizer Autoencoder (VQ-VAE) \cite{van2017neural} instead of using a publicly available pretrained models. The images are encoded into a latent compact representation using the VQ-VAE's encoder for the DM, and its output is then reconstructed by the VQ-VAE's decoder. 
	We don't change the training process and the pruned dataset is obtained in a pretraining phase either with or without additional computation dependending on the pruning method. During training, we use a batch size of 256 at resolution 256x256 and run for 20 hours using 4 NVIDIA H100 GPUs on ImageNet (a total of 120k iterations). During training on Imagenet, the DM is conditioned on the image class label. On CelebA-HQ, we train an unncoditional model using weaker gpus with 32 batch size and for less hours (a total of 220k iterations). 
	\subsection{Evaluation over CelebA-HQ}
\begin{figure}
		\centering
		\captionsetup{justification=centering}
		\begin{subfigure}[]{0.47\textwidth}
			\includegraphics[width=1\linewidth]{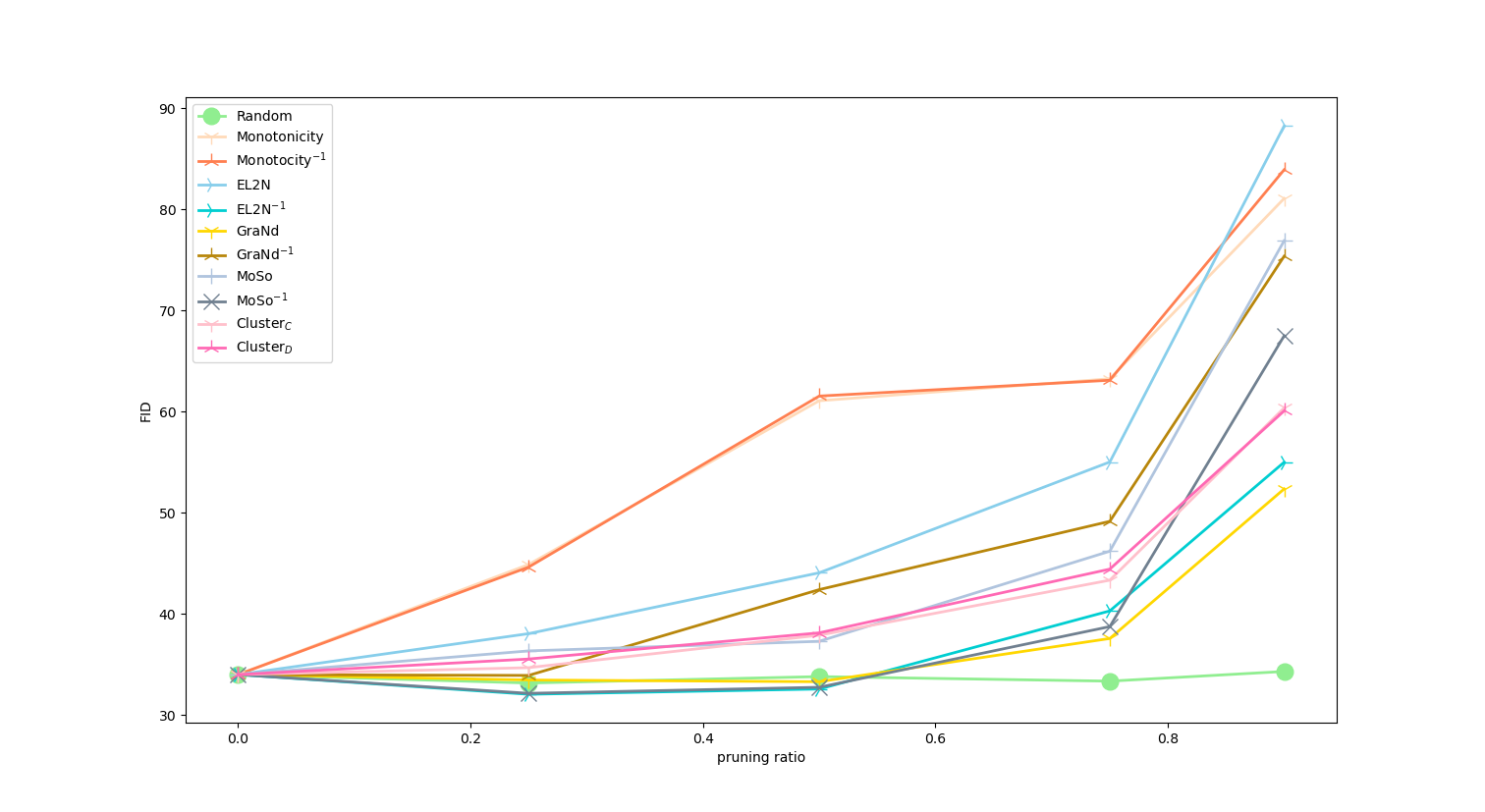}
			\vspace{-2em}
			\caption{Evaluation over CelebA-HQ using 4k samples.}
			\vspace{-0.3em}
			\label{fig:fid_celebhq}
		\end{subfigure}
		\begin{subfigure}[]{0.47\textwidth}
			\includegraphics[width=1\linewidth]{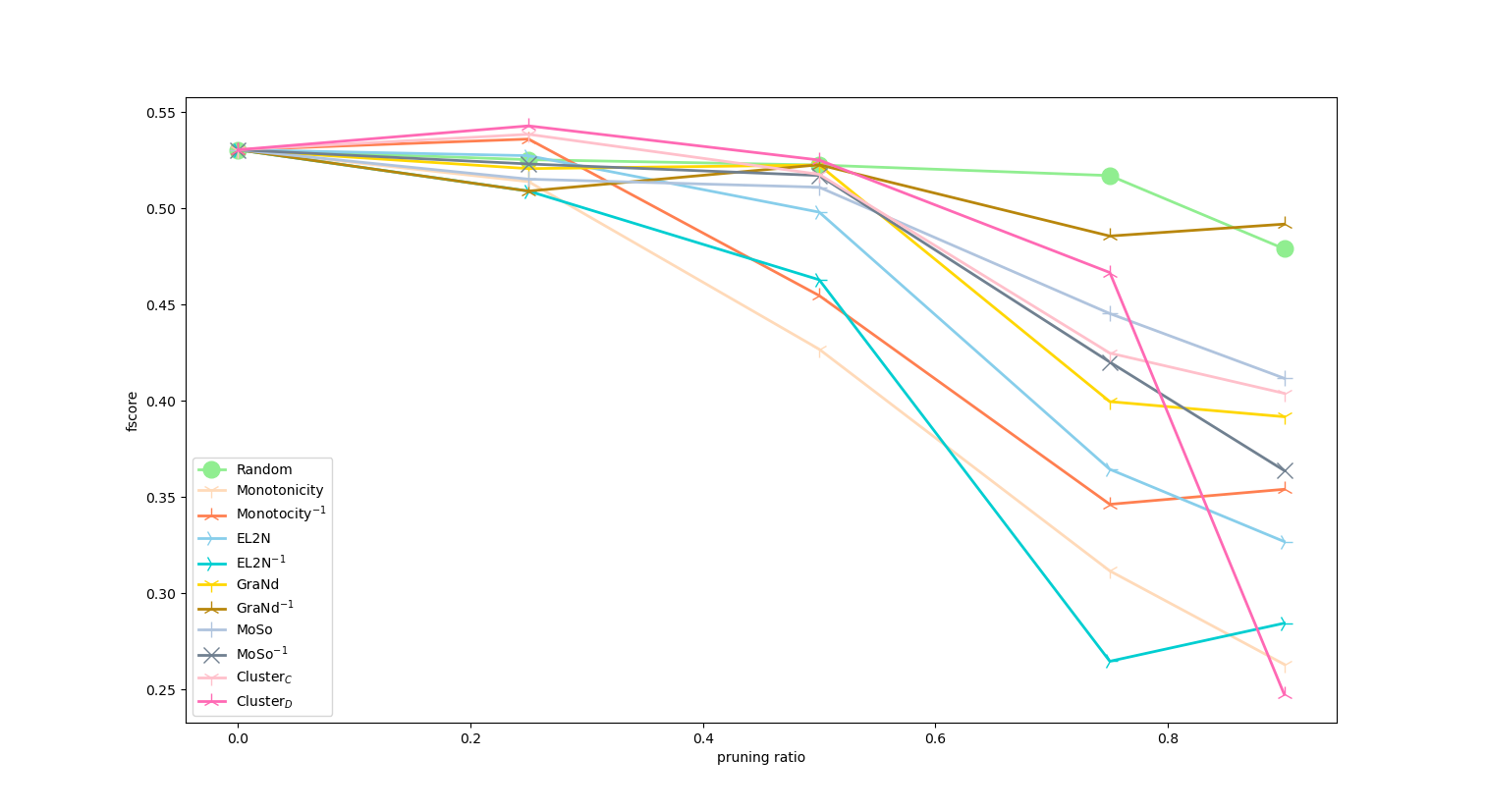}
			\vspace{-2em}
			\caption{F-score evaluation over CelebA-HQ using 4k samples.}
			\vspace{-0.3em}
			\label{fig:fscore_celebhq}
		\end{subfigure}
		
		\begin{subfigure}[]{0.47\textwidth}
			\includegraphics[width=1\linewidth]{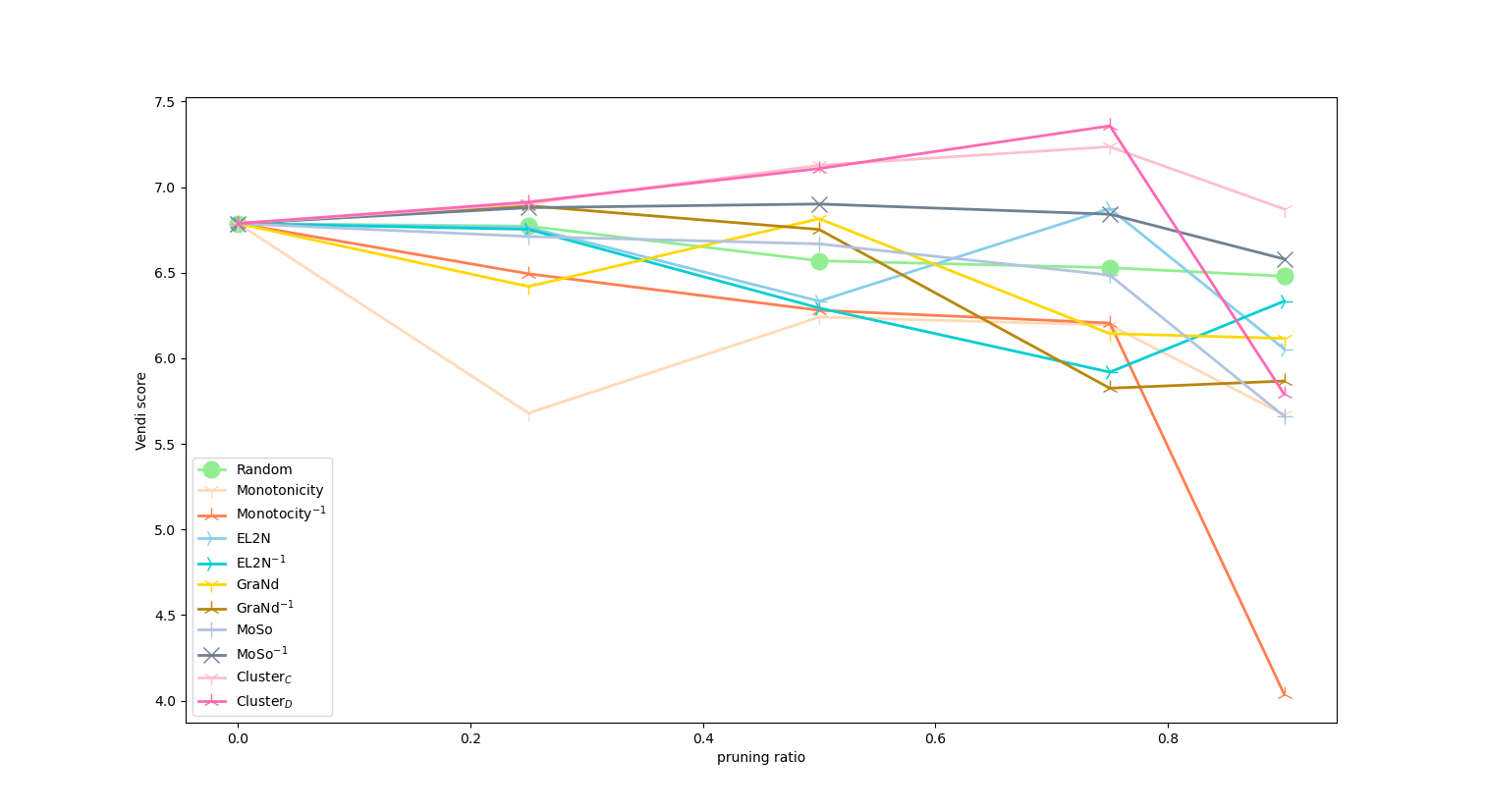}
			\vspace{-2em}
			\caption{Vendi score over CelebA-HQ.}
			\vspace{-0.3em}
			\label{fig:vendi_celebhq}
		\end{subfigure}
	\end{figure}

	\myparagraph{Quantitative metrics.}
	Figure~\ref{fig:fid_celebhq} shows the FID curve for different strategies at different pruning ratios (PR). We observe that certain pruning strategies such as Monotonicity, its inverse Monotonicity$^{-1}$ and EL2N perform worse proportionally to the pruning ratio. In fact, for these two methods the deterioration is strikingly more drastic than the other methods. \textit{Monotonicity} and its inverse perform on par with each other indicating the method has no significance when learning a distribution. EL2N deteriorates the most when PR=0.9. There is a huge gap between EL2N and its inverse (in favor of its inverse), in agreement with our earlier reasoning that lower loss is indicative of high density regions. MoSo $^{-1}$ also peforms better indicating it is ineffective, contradicting its efficacy in discrminative models. The curves of MoSo$^{-1}$ and EL2N$^{-1}$  are slightly below \textit{random} for $PR\leq 0.5$, after that their curves increase proportionally to the pruning ratio. It is also interesting that for $PR\leq0.5$ there is a slight improvement of these 2 strategies and the random baseline compared to no pruning. This indicates that DMs are such powerful generative models capable of extrapolating to the whole distribution given only a portion of the dataset. The curve of GraNd is consistely below its inverse curve indicating that this strategy is meaningful in terms of pruning otpimization, which is consistent with larger gradients being indicative of higher probability density. However, we observe that the curve of the random baseline is consistently good for all PR values, while all other strategies deteriorate significantly when 90\% of the data is pruned, i.e. the model is trained using only 3k samples. This indicates that at a high PR it is not worthwhile to apply a pruning strategy that requires computation and it is better to select samples randomly, which will cover the entire distribution.
	Figure~\ref{fig:fscore_celebhq} shows the F-score curves across different PRs. For $PR\leq0.5$, the clustering strategy yields the highest scores, then only the random baseline retains its performance, where at $PR>0.75$ there is a more noticeable drop, unlike its FID curve which was maintained across all PRs. For some strategies such as MoSo$^{-1}$ and GraNd, the curve behaves in accordance with the FID curve for a smaller range of PRs, for example $PR\leq0.5$, however, the curves of EL2N and its inverse behave in a manner contrary to their FID and F-score plots.   
	In figure~\ref{fig:vendi_celebhq} we plot the Vendi score  across the different PRs. For most methods, the behavior is similar to the FID curve, but for EL2N for example the deterioration in Vendi score is less substantial than in FID which indicates that even though the output is generating diverse outputs, their quality is low. 
	In table~\ref{tab:eval_overfitting}, we examine whether the model memorizes the training samples especially as the pruning ration increases substantially. To that end, we extract the fatures using CLIP visual model and compute average distance between the generated samples and their match. However, these values indicate that there is no memorization on average, although by looking at the distance of some samples, we observe that the model replicates training samples, and where the distance would then drop to $\approx 3$. In fact, at $PR=0.95$ the average minimum distance is $\approx 3$, indicating that pruning any more than 90\%  breaks the model's capability of generating novel samples from the distribution. We also observed that the random baseline on CelebA-HQ has the highest tolerance threshold before it starts memorizing the samples.
	
\begin{figure}[]
		\setlength\tabcolsep{3pt}
		\resizebox{0.5\textwidth}{!}{%
			
			%lllll
			\begin{tabular}{ccccc}

				\multicolumn{1}{c}{Random}{\includegraphics[height=2cm, width=2cm]{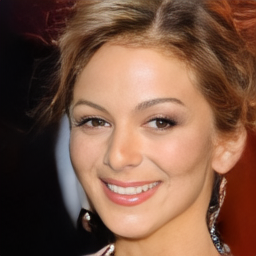}}& {\includegraphics[height=2cm, width=2cm]{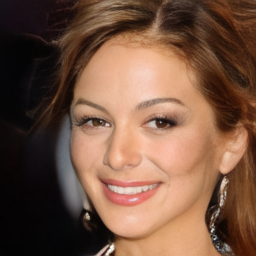}}&
				{\includegraphics[height=2cm, width=2cm]{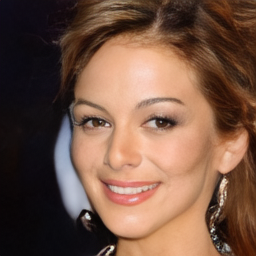}}&
				{\includegraphics[height=2cm, width=2cm]{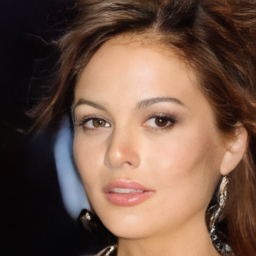}}&
				{\includegraphics[height=2cm, width=2cm]{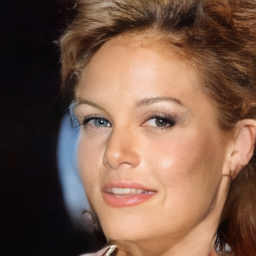}}%&{\includegraphics[height=2cm, width=2cm]{images/celebhq_nearest_match/random}}
				\\

				\multicolumn{1}{c}{Monoticity}    {\includegraphics[height=2cm, width=2cm]{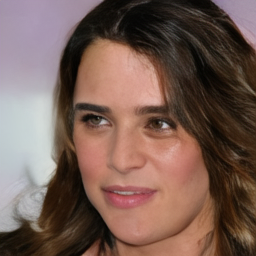}}& {\includegraphics[height=2cm, width=2cm]{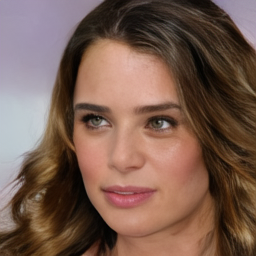}}&
				{\includegraphics[height=2cm, width=2cm]{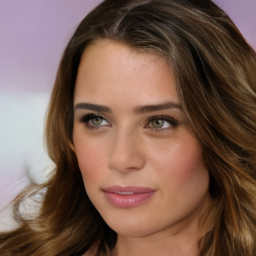}}&
				{\includegraphics[height=2cm, width=2cm]{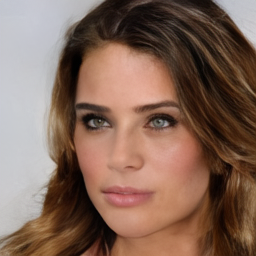}}&
				{\includegraphics[height=2cm, width=2cm]{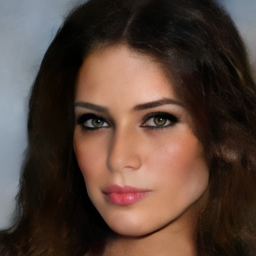}}%&{\includegraphics[height=2cm, width=2cm]{images/celebhq_nearest_match/forgetting}}
				\\
				
				\multicolumn{1}{c}{Monoticity$^{-1}$} {\includegraphics[height=2cm, width=2cm]{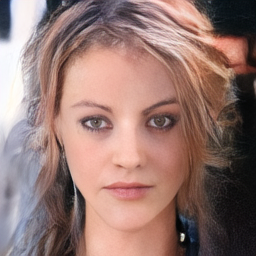}}&   {\includegraphics[height=2cm, width=2cm]{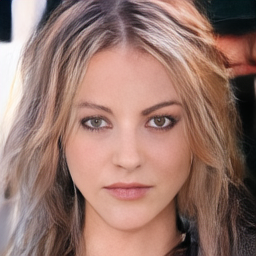}}&
				{\includegraphics[height=2cm, width=2cm]{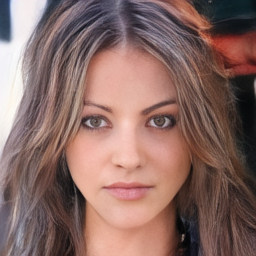}}&
				{\includegraphics[height=2cm, width=2cm]{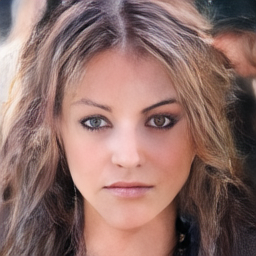}}&
				{\includegraphics[height=2cm, width=2cm]{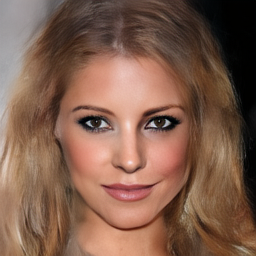}}

				\\
				\multicolumn{1}{c}{GraNd} {\includegraphics[height=2cm, width=2cm]{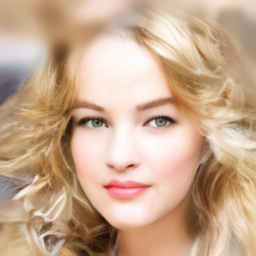}}&   {\includegraphics[height=2cm, width=2cm]{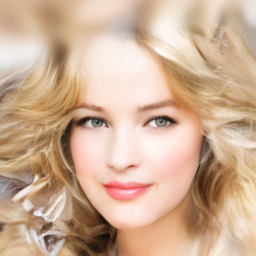}}&
				{\includegraphics[height=2cm, width=2cm]{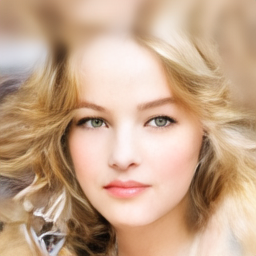}}&
				{\includegraphics[height=2cm, width=2cm]{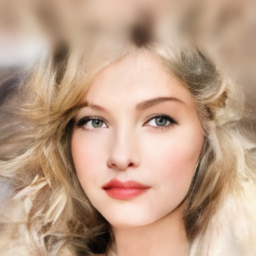}}&
				{\includegraphics[height=2cm, width=2cm]{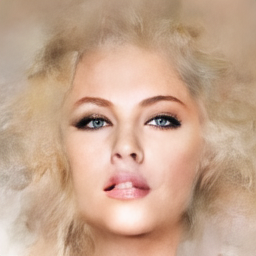}}
				\\

				\multicolumn{1}{c}{GraNd $^{-1}$}   {\includegraphics[height=2cm, width=2cm]{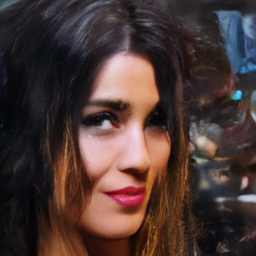}}& {\includegraphics[height=2cm, width=2cm]{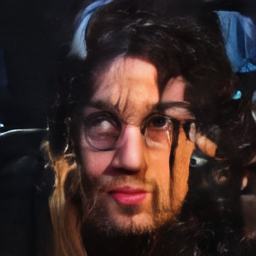}}&
				{\includegraphics[height=2cm, width=2cm]{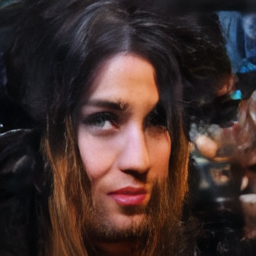}}&
				{\includegraphics[height=2cm, width=2cm]{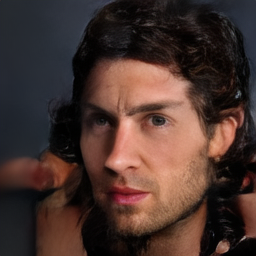}}&
				{\includegraphics[height=2cm, width=2cm]{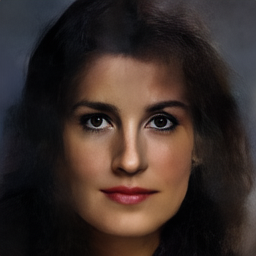}}
				\\
				
				\multicolumn{1}{c}{EL2N}   {\includegraphics[height=2cm, width=2cm]{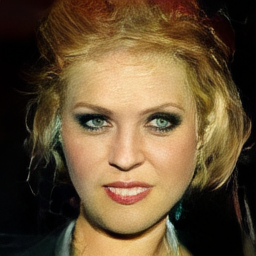}}&   {\includegraphics[height=2cm, width=2cm]{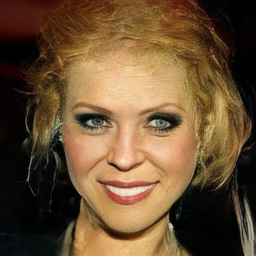}}&
				{\includegraphics[height=2cm, width=2cm]{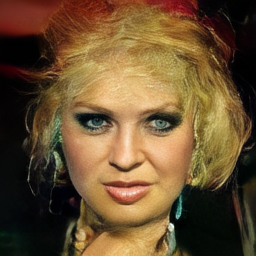}}&
				{\includegraphics[height=2cm, width=2cm]{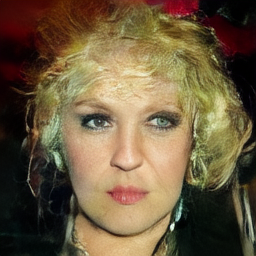}}&
				{\includegraphics[height=2cm, width=2cm]{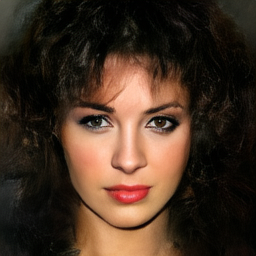}}
				\\

				\multicolumn{1}{c}{EL2N $^{-1}$}   {\includegraphics[height=2cm, width=2cm]{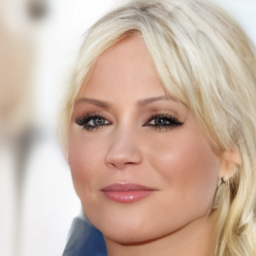}}& {\includegraphics[height=2cm, width=2cm]{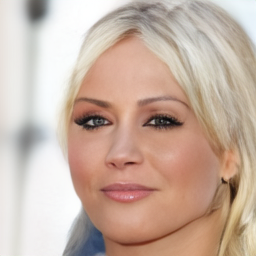}}&
				{\includegraphics[height=2cm, width=2cm]{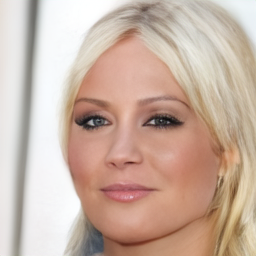}}&
				{\includegraphics[height=2cm, width=2cm]{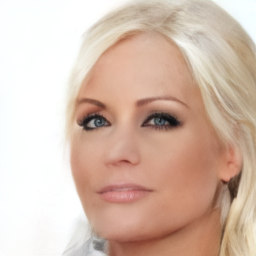}}&
				{\includegraphics[height=2cm, width=2cm]{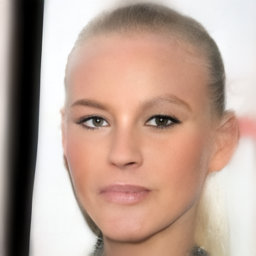}}
				%&{\includegraphics[height=2cm, width=2cm]{images/celebhq_nearest_match/loss_inv}}
				\\

				\multicolumn{1}{c}{MoSo}  {\includegraphics[height=2cm, width=2cm]{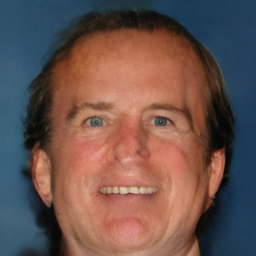}}& {\includegraphics[height=2cm, width=2cm]{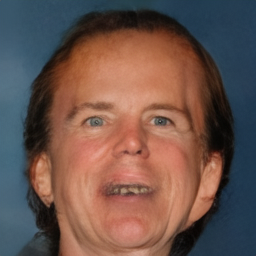}}&
				{\includegraphics[height=2cm, width=2cm]{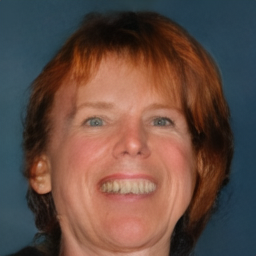}}&
				{\includegraphics[height=2cm, width=2cm]{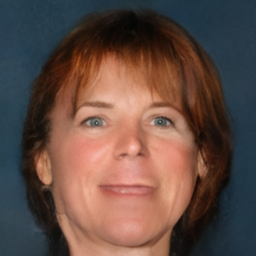}}&
				{\includegraphics[height=2cm, width=2cm]{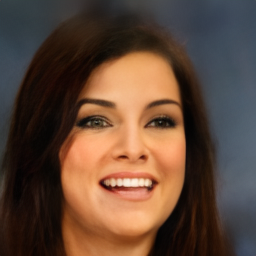}}
				\\
				
				\multicolumn{1}{c}{MoSo $^{-1}$} {\includegraphics[height=2cm, width=2cm]{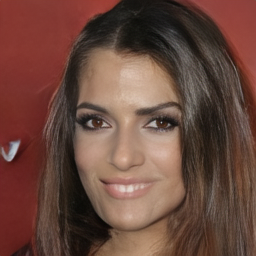}}&  {\includegraphics[height=2cm, width=2cm]{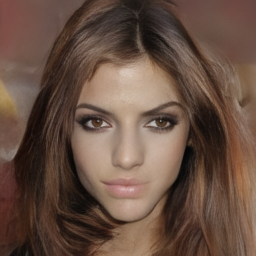}}&
				{\includegraphics[height=2cm, width=2cm]{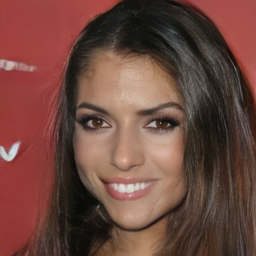}}&
				{\includegraphics[height=2cm, width=2cm]{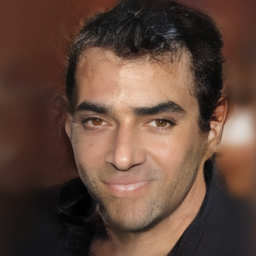}}&
				{\includegraphics[height=2cm, width=2cm]{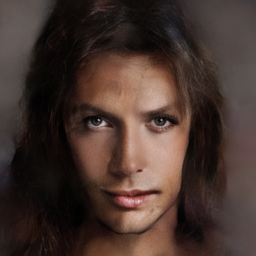}}
				\\
				
				\multicolumn{1}{c}{Cluster$_C$}  {\includegraphics[height=2cm, width=2cm]{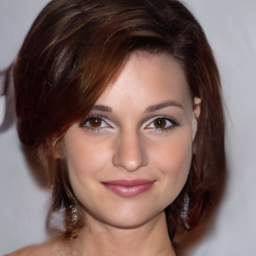}}& {\includegraphics[height=2cm, width=2cm]{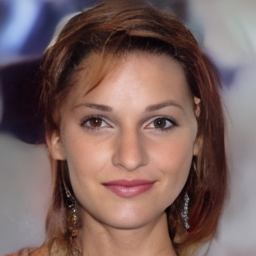}}&
				{\includegraphics[height=2cm, width=2cm]{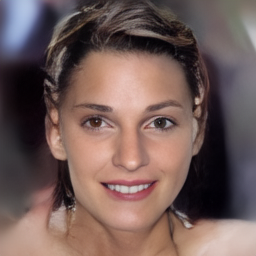}}&
				{\includegraphics[height=2cm, width=2cm]{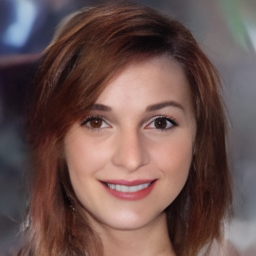}}&
				{\includegraphics[height=2cm, width=2cm]{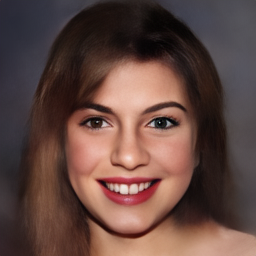}}
				
				\\

				\multicolumn{1}{c}{Cluster$_D$}{\includegraphics[height=2cm, width=2cm]{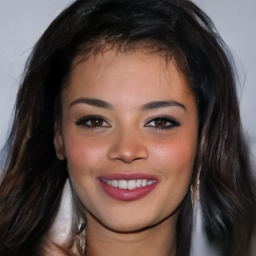}}&   {\includegraphics[height=2cm, width=2cm]{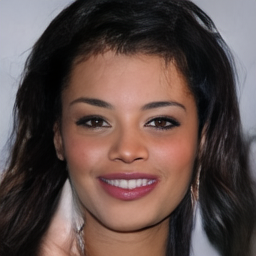}}&
				{\includegraphics[height=2cm, width=2cm]{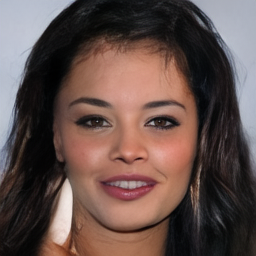}}&
				{\includegraphics[height=2cm, width=2cm]{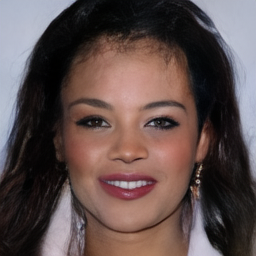}}&
				{\includegraphics[height=2cm, width=2cm]{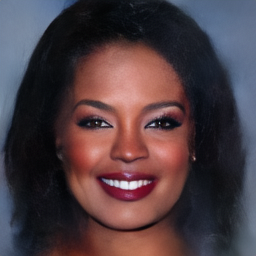}}
				
				\\
				
				\multicolumn{1}{c}{unpruned}& \multicolumn{1}{c}{PR=0.25}&  \multicolumn{1}{c}{0.5}&  \multicolumn{1}{c}{0.75}&  \multicolumn{1}{c}{0.9}
				
		\end{tabular}}
		\vspace{-1em}
		\caption{Qualitative results generated for different PRs for pruning methods and their inverse while varying PR.}
		\vspace{-1em}
		\label{fig:qual_celebhq}
	\end{figure}

	\textbf{Qualitative evaluation.}
	Figure \ref{fig:qual_celebhq} shows qualititative results for each pruning method at different PRs. For each method, the initial noise is kept fixed so we can look at the changes as we vary the PR. The random baseline exhibits little changes across the different PRs, where the most visible feature change occurs at PR=0.75 as it goes from a smiling to non-smiling expression and the left eye colors becomes darker than the right eye, but there is no perceived degradation in the quality. When PR=0.9, however, there is a perceived degradation in the lighting, the chin area, and the left eye color becomes different from the right eye. The beheviour correlates well with the FID curve. In the monoticity strategy, we don't observe a clear degradation, but at PR=0.9 the model replicates a training sample. For its inverse Monoticity $^{-1}$, we notice a visible improvement at when PR=0.25 and 0.5 compared to no pruning, but then the quality degrades visibly when PR=0.75 and at PR=0.9 the sample is replicated. In GraNd there is no perceived degradation in quality but its inverse shows clear quality degradation for all PRs, which correlates well with its FID curve as well. This seems to indicate that while GraNd is not the best strategy to prune samples, using its inverse to select unimportant samples in terms of their gradient is very detrimental to the model. EL2N also demonstrates clear quality degradation already when PR=0.25, which is also consistent with FID curve. In MoSo, there is a clear degradation when PR=0.25 and 0.5, although there is an improvement at PR=0.75. But this particular initial noise was challenging for most of the pruning methods since even with the unpruned dataset, the model shows distortion. MoSo inverse shows clear degradation when PR=0.75 and 0.9. Cluster$_{C}$ shows some variation while Cluster$_{D}$ shows little variations across the different PRs.
	To contrast the sampling of the different pruning methods using the same initial noise, we include further qualitative samples in the supplementary material.  
	
	\begin{table}[h!]
		
		\resizebox{0.5\textwidth}{!}{
			
			\begin{tabular}{lllllll}
				
				\hline
				PR& Random & Monotonicity &GraNd&EL2N& MoSo& cluster$_C$\\
				&  & Monotonicity$^{-1}$ & GraNd$^{-1}$ & EL2N$^{-1}$& MoSo$^{-1}$& cluster$_D$ \\
				\hline
				\multirow{2}{*}{0.25} &\multirow{2}{*}{5.6889} & 5.5908& 5.6703& 5.7179&5.6418&7.7434
				\\
				
				&& 5.5829 & 5.7638 &5.7036&5.7132&5.8241\\

				\multirow{2}{*}{0.5} & \multirow{2}{*}{5.7667}  &\graycolor{20} {5.4036}&\graycolor{20} 5.6512 & \graycolor{20} 5.7792 &\graycolor{20} 5.5885&\graycolor{20} 5.9830
				\\

				&& \graycolor{20} 5.3889 &\graycolor{20}  5.6864 & \graycolor{20} 5.6389&\graycolor{20} 5.7085&\graycolor{20} 5.9657\\
				
				\multirow{2}{*}{0.75}  &\multirow{2}{*}{5.7034} &5.4137& 5.6699& 5.6139 &5.4828&6.0654
				\\
				
				&&5.4188 & 5.6150& 5.6139&5.7324&5.9814
				\\

				\multirow{2}{*}{0.9}& \multirow{2}{*}{5.5979}  & \graycolor{20}5.5681&\graycolor{20} 5.4923 & \graycolor{20}6.0452 &\graycolor{20}5.2547&\graycolor{20}5.7440
				\\
				&&\graycolor{20}5.2165 &\graycolor{20}5.4219&\graycolor{20}5.0786&\graycolor{20}5.7092&\graycolor{20}5.8668\\
				\hline
				Unpruned &5.7783\\
				Validation &4.4972\\
				\hline
			\end{tabular}
		}
		
		\vspace{-0.5em}
		
		\caption{The average minimum distance for generated images to the training set of CelebA-HQ.}
		\vspace{-1.5em}
		\label{tab:eval_overfitting}
		
	\end{table} 
	
	\subsection{Evaluation over ImageNet}
\begin{figure}
			\centering
			\captionsetup{justification=centering}
			\begin{subfigure}[]{0.5\textwidth}
				\includegraphics[width=1\linewidth]{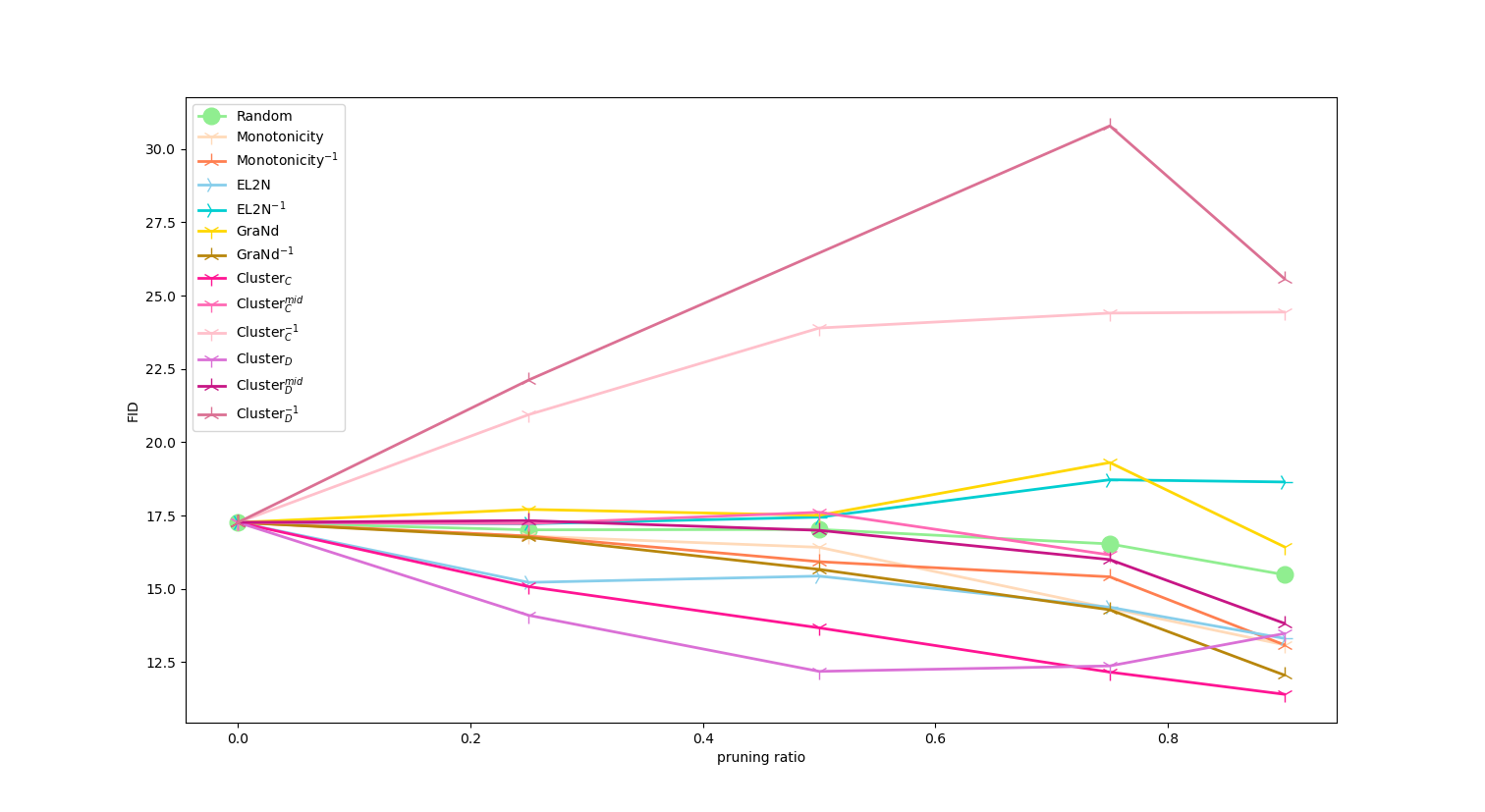}
				\vspace{-2.2em}
				\caption{FID over ImageNet. }
				\vspace{-0.3em}
				\label{fig:fid_imagenet}
			\end{subfigure}
			\begin{subfigure}[]{0.5\textwidth}
				\includegraphics[width=1\linewidth]{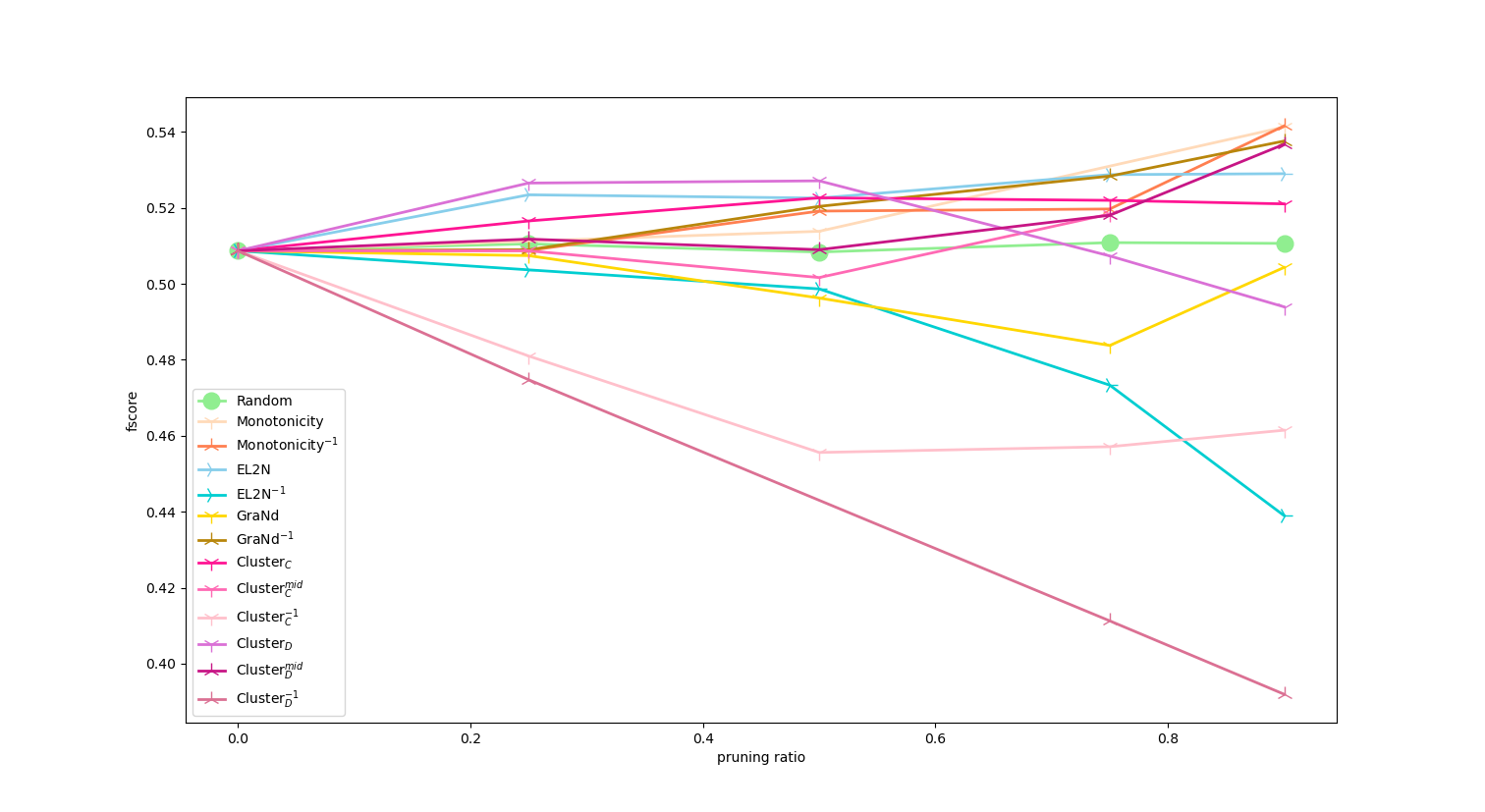}
				\vspace{-2em}
				\caption{F-score over ImageNet.}
				\vspace{-0.3em}
				\label{fig:fscore_imagenet}
			\end{subfigure}
			
			\begin{subfigure}[]{0.5\textwidth}
				\includegraphics[width=1\linewidth]{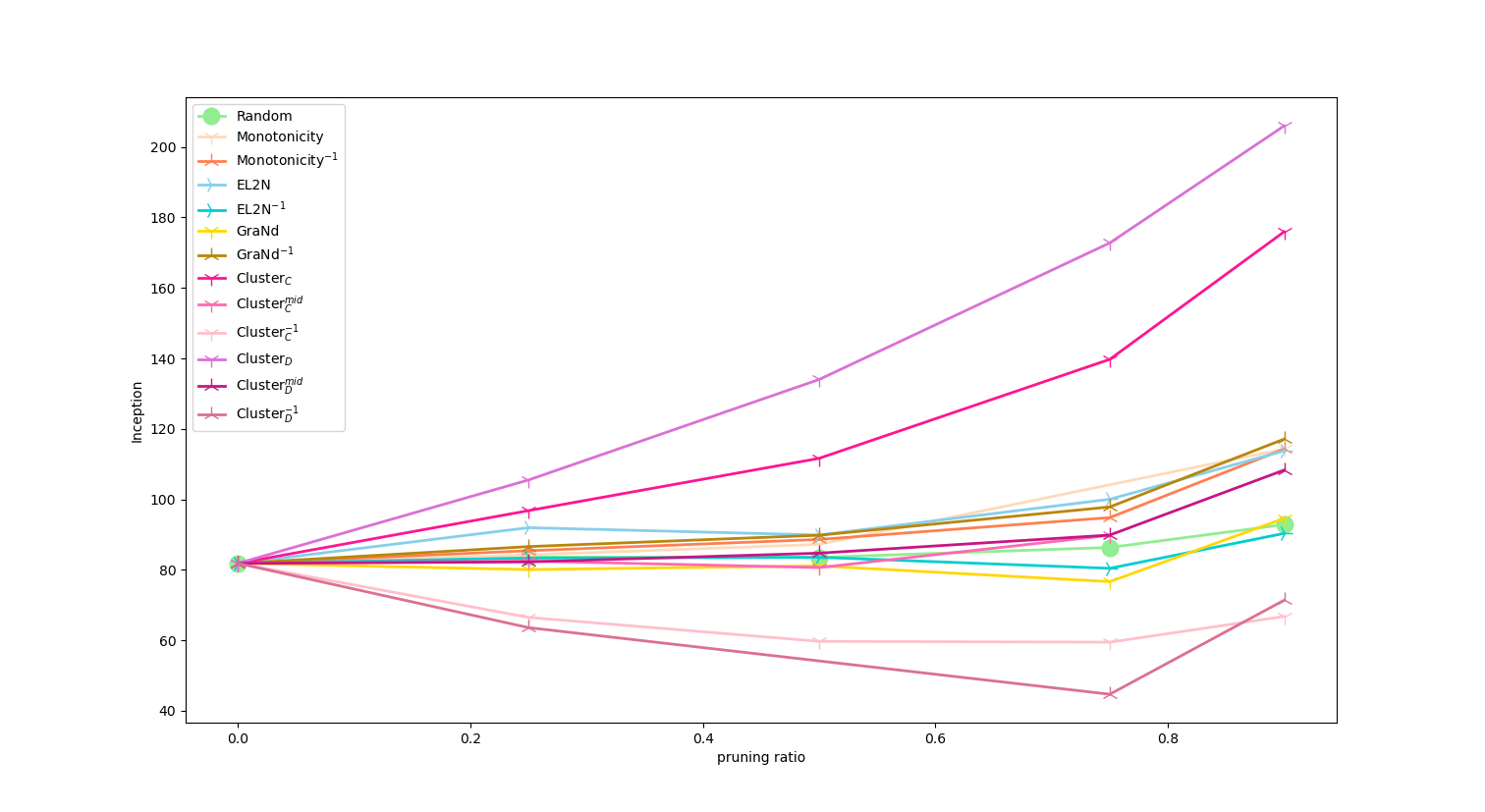}
				\vspace{-2.2em}
				\caption{Inception score over ImageNet.}
				\vspace{-0.3em}
				\label{fig:inception_imagenet}
			\end{subfigure}

			\begin{subfigure}[]{0.5\textwidth}
				\includegraphics[width=1\linewidth]{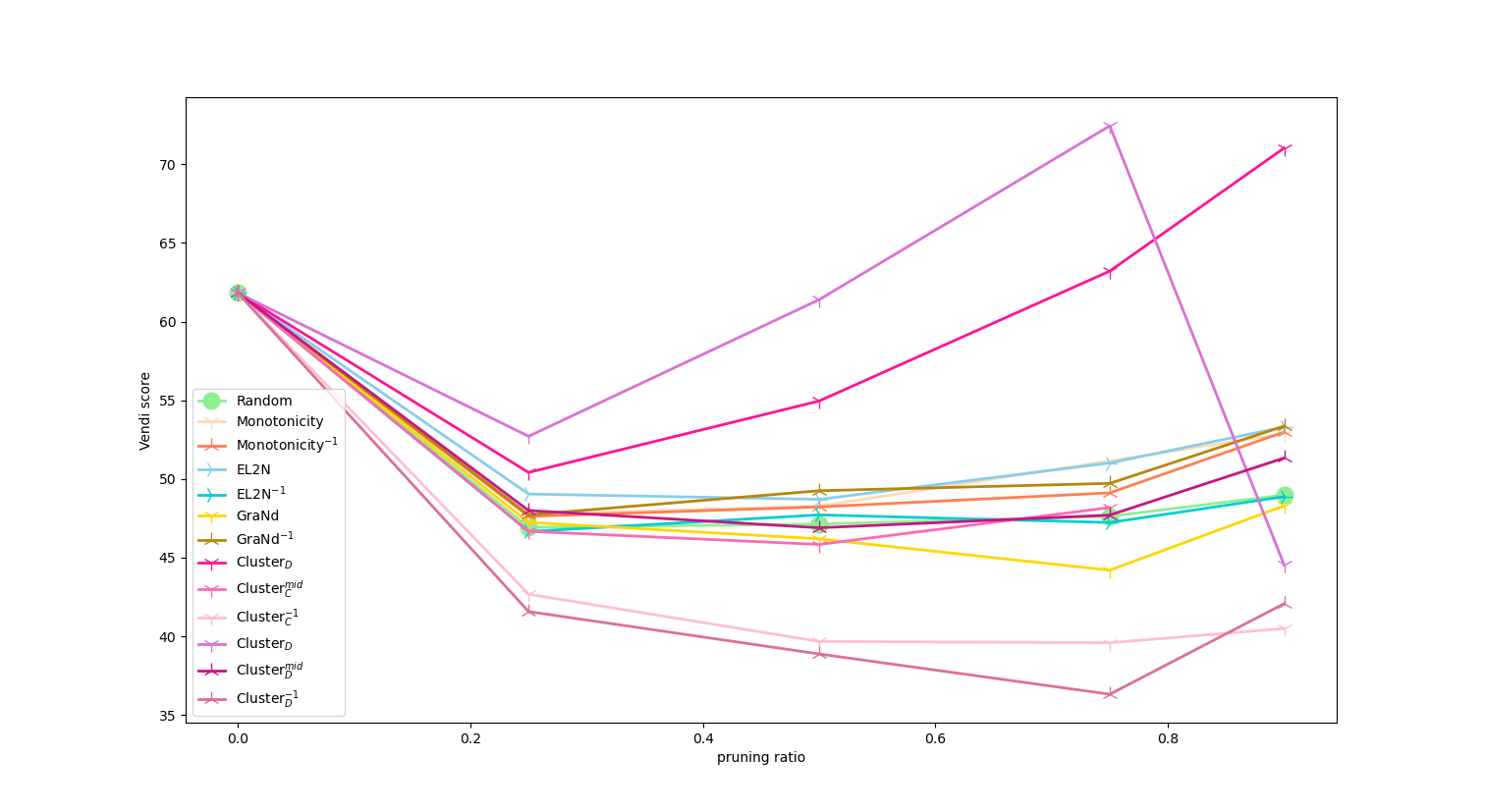}
				\vspace{-2.2em}
				\caption{Vendi score over ImageNet.}
				\vspace{-0.5em}
				\label{fig:vendi_imagenet}
			\end{subfigure}
			
		\end{figure}

	\myparagraph{Quantitative metrics.}
	On ImageNet, we observe variation in the behaviour patterns compared to CelebA-HQ. In figure~\ref{fig:fid_imagenet} , we notice for example that nearly all pruning strategies were beneficial for training. In fact, we see that the curves are monotonically decreasing as we increase the pruning ratio with some degradation for GraNd and EL2N$^{-1}$ for $PR>0.5$ and a drastic degrdation when applying Cluster$_{C/D}^{-1}$. Across all PRs, Cluster$_{C/D}$ yields the lowest FID and using DINO features outperforms CLIP features  except for $PR>0.75$ where Cluster$_{D}$'s performance drops. At $PR=0.9$ the gap between unpruned and Cluster$_C$ is $\approx 5$. Another clear different pattern is that GraNd$^{-1}$ performs better than GraNd as opposed to CelebA-HQ, and EL2N performing worse than its inverse, possibly indicating the presence of outliers, . Also most pruning strategies including their inverse outperform the random baseline.
	In figure~\ref{fig:fscore_imagenet} where we evaluate the F-score, we see that clustering curves behaves in a consistent manner to their FID curves. EL2N's F-score curve behaves inversely to its FID curve for $PR>0.5$ where the decrease results from a decrease in precision (fidelity). We attach the separate precision and recall curves in the supplementary. In figure~\ref{fig:inception_imagenet}, the curves behave in a similar manner to the FID curves with slight fluctuations. In figure~\ref{fig:vendi_imagenet}, cluster$_{D}$'s Vendi score drops more markedly than its FID at $PR>0.75$, inndicating that despite the sample high fidelity (as evident from the FID and inception scores), the diversity is noticeably degraded.  
	We don't provide the average minimum distance for ImageNet as comparing the validation set and generated sets against 1 million samples to find their match is implausible. We observed the sample memorization phenonmen at very high pruning ratios such as $PR=0.99$ when we inspected a few samples. 
	
\begin{figure}[]
		\setlength\tabcolsep{3pt}
		\resizebox{0.5\textwidth}{!}{%
			%llllllllll
			
			\begin{tabular}{cccccccccc}
				
				\multicolumn{1}{c}{Random}{\includegraphics[height=2cm, width=2cm]{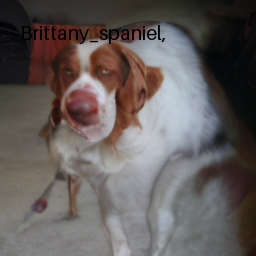}}& 
				{\includegraphics[height=2cm, width=2cm]{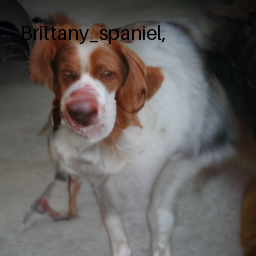}}&		
				{\includegraphics[height=2cm, width=2cm]{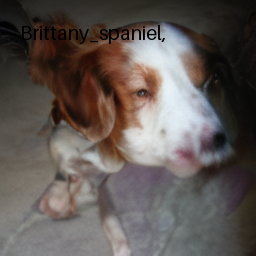  }}
				&
				{\includegraphics[height=2cm, width=2cm]{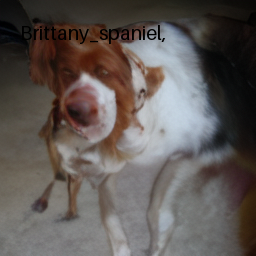 }}
				&
				
				{\includegraphics[height=2cm, width=2cm]{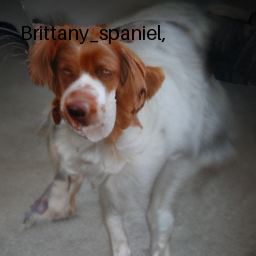}}
				
				\\
				
				\multicolumn{1}{c}{Monotonicity}{\includegraphics[height=2cm, width=2cm]{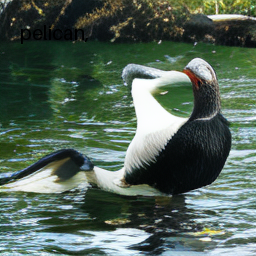}}& 
				{\includegraphics[height=2cm, width=2cm]{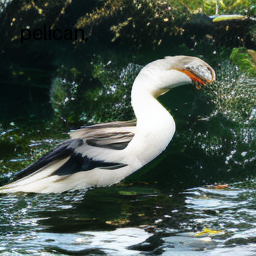}}&		
				{\includegraphics[height=2cm, width=2cm]{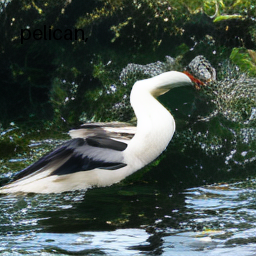  }}
				&
				{\includegraphics[height=2cm, width=2cm]{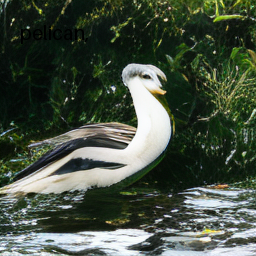}}
				& 
				
				{\includegraphics[height=2cm, width=2cm]{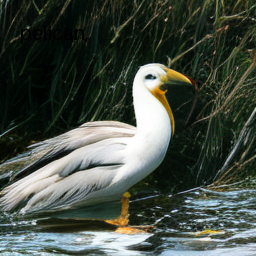 }}

				\\

				\multicolumn{1}{c}{Monotonicity $^{-1}$}{\includegraphics[height=2cm, width=2cm]{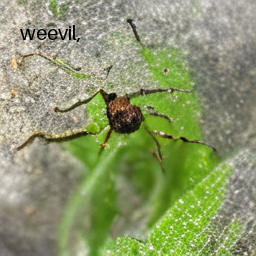}}& 
				{\includegraphics[height=2cm, width=2cm]{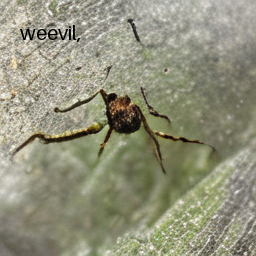}}&		
				{\includegraphics[height=2cm, width=2cm]{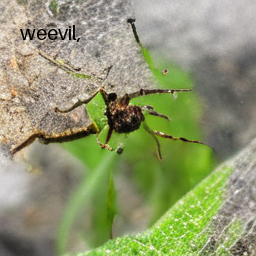}}
				&
				{\includegraphics[height=2cm, width=2cm]{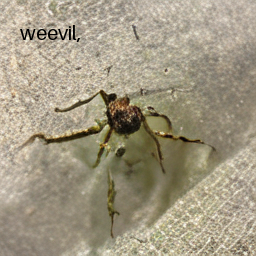}}
				&
				
				{\includegraphics[height=2cm, width=2cm]{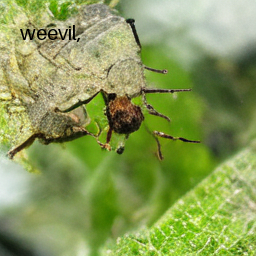}}
				
				\\
				
				\multicolumn{1}{c}{GraNd}{\includegraphics[height=2cm, width=2cm]{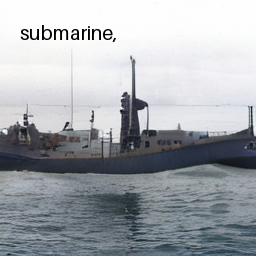}}& 
				{\includegraphics[height=2cm, width=2cm]{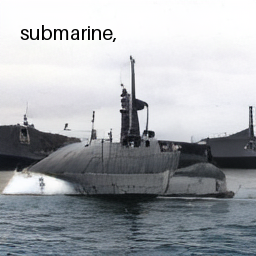}}&		
				{\includegraphics[height=2cm, width=2cm]{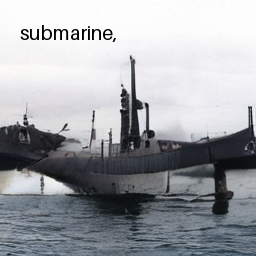 }}
				&
				{\includegraphics[height=2cm, width=2cm]{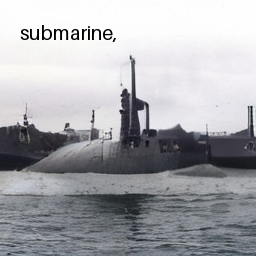}}
				& 
				
				{\includegraphics[height=2cm, width=2cm]{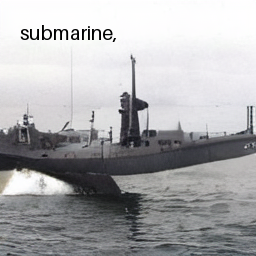}}
				
				\\

				\multicolumn{1}{c}{GraNd $^{-1}$}{\includegraphics[height=2cm, width=2cm]{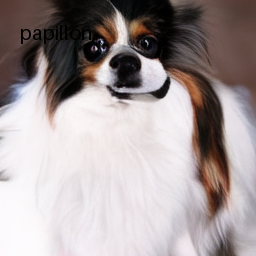}}& 
				{\includegraphics[height=2cm, width=2cm]{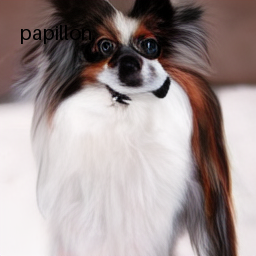}}&		
				{\includegraphics[height=2cm, width=2cm]{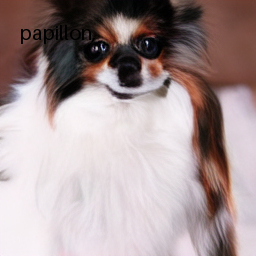}}
				&
				{\includegraphics[height=2cm, width=2cm]{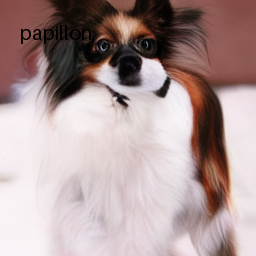}}
				& 
				
				{\includegraphics[height=2cm, width=2cm]{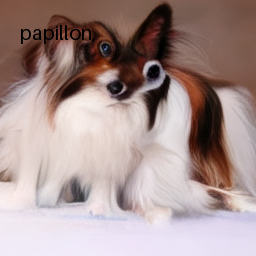}}
				
				\\

				\multicolumn{1}{c}{EL2N}   {\includegraphics[height=2cm, width=2cm]{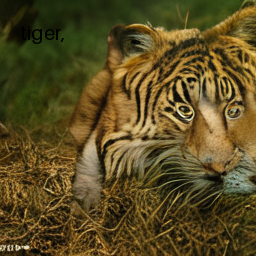}}&
				{\includegraphics[height=2cm, width=2cm]{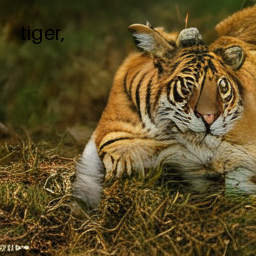}} &
				
				{\includegraphics[height=2cm, width=2cm]{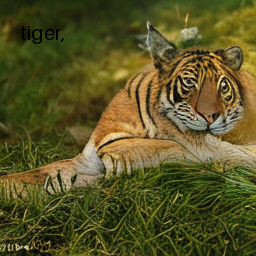}}
				&
				{\includegraphics[height=2cm, width=2cm]{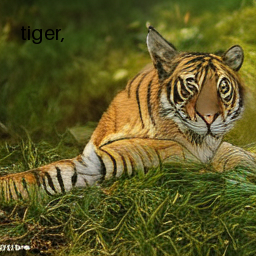}}
				
				&
				{\includegraphics[height=2cm, width=2cm]{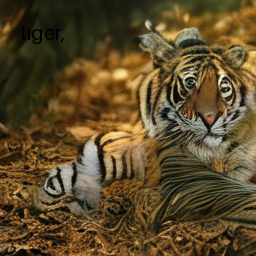}} \\
				
				\multicolumn{1}{c}{EL2N $^{-1}$}   {\includegraphics[height=2cm, width=2cm]{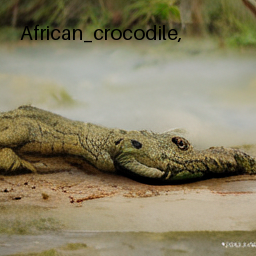}}&
				{\includegraphics[height=2cm, width=2cm]{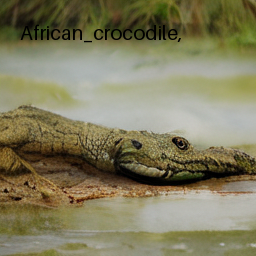}} &
				
				{\includegraphics[height=2cm, width=2cm]{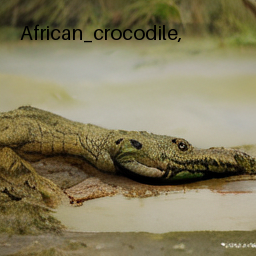}}
				&
				{\includegraphics[height=2cm, width=2cm]{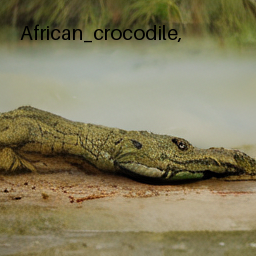}}
				
				&
				{\includegraphics[height=2cm, width=2cm]{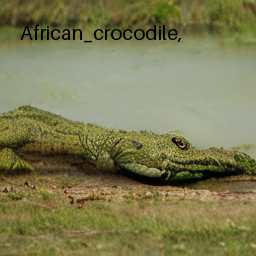}}
				
				\\
				
				\multicolumn{1}{c}{Cluster$_C$}  {\includegraphics[height=2cm, width=2cm]{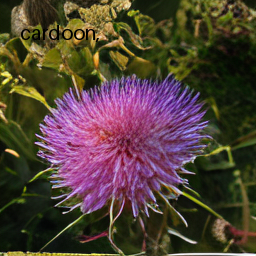}}&
				{\includegraphics[height=2cm, width=2cm]{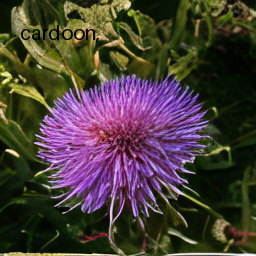}}&
				{\includegraphics[height=2cm, width=2cm]{images/imagenet/00059_061-DiT-XL-2_conditional_imagenet_vaeretrained_fm_inversecluster_clip_pruned05.png}}
				&
				{\includegraphics[height=2cm, width=2cm]{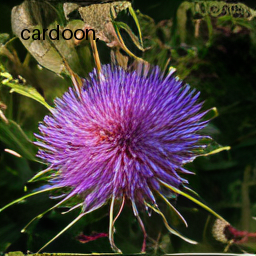}}
				&	{\includegraphics[height=2cm, width=2cm]{images/imagenet/00059_061-DiT-XL-2_conditional_imagenet_vaeretrained_fm_inversecluster_clip_pruned05.png}}
				\\

				\multicolumn{1}{c}{Cluster$_C^{-1}$}{\includegraphics[height=2cm, width=2cm]{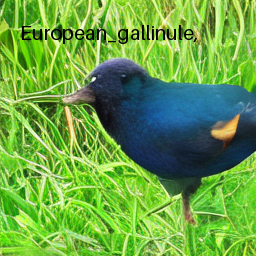}}&
				{\includegraphics[height=2cm, width=2cm]{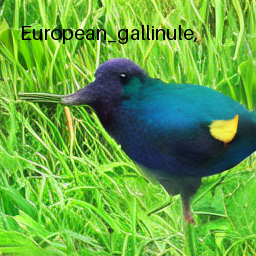}}&
				{\includegraphics[height=2cm, width=2cm]{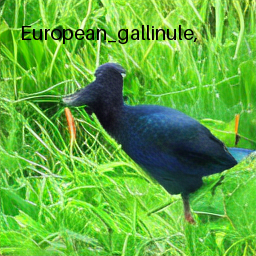}}&
				{\includegraphics[height=2cm, width=2cm]{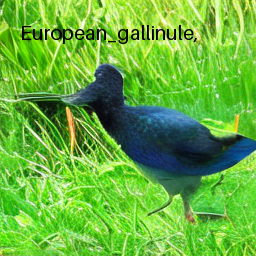}}&
				{\includegraphics[height=2cm, width=2cm]{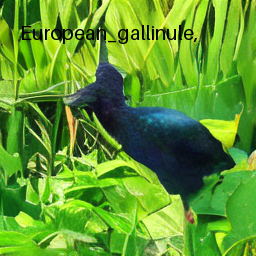}}
				\\
				\multicolumn{1}{c}{Cluster$_D$}{\includegraphics[height=2cm, width=2cm]{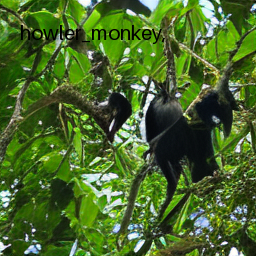}}&
				{\includegraphics[height=2cm, width=2cm]{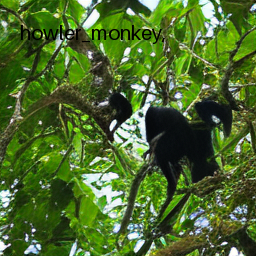}}&
				{\includegraphics[height=2cm, width=2cm]{images/imagenet/00050_063-DiT-XL-2_conditional_imagenet_vaeretrained_fm_inversecluster_dino_pruned025}}&
				{\includegraphics[height=2cm, width=2cm]{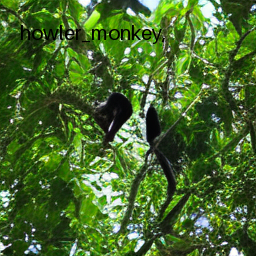}}&
				{\includegraphics[height=2cm, width=2cm]{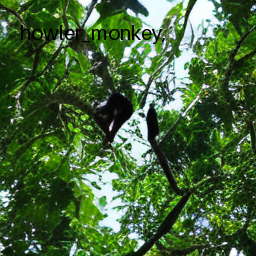}}
				
				\\
				
				\multicolumn{1}{c}{unpruned}& \multicolumn{1}{c}{PR=0.25}& \multicolumn{1}{c}{PR=0.5}& \multicolumn{1}{c}{PR=0.75}& \multicolumn{1}{c}{PR=0.9}
				
		\end{tabular}}
		\vspace{-1em}
		\caption{Qualitative results that display samples while varying PR. Notice that the quality does not degrade as PR increases, in agreement with the FID curve.}
		\vspace{-1em}
		\label{fig:qual_imagenet}
	\end{figure}

\begin{figure}[]

		\hfill
		%parbox{\dimexpr\linewidth-0.1em}{%
			\setlength\tabcolsep{3pt}
			
			\resizebox{0.45\textwidth}{!}{%

				\begin{tabular}{ccc}
					
					\multicolumn{1}{c}{CLIP}	{\includegraphics[height=2cm, width=2cm]{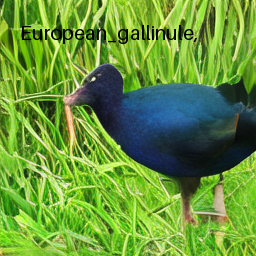}}&
					{\includegraphics[height=2cm, width=2cm]{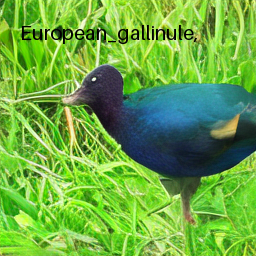}}&
					
					{\includegraphics[height=2cm, width=2cm]{images/imagenet/00085_009-DiT-XL-2_conditional_imagenet_vaeretrained_fm_cluster_clip_pruned05}}
					
					\\
					\multicolumn{1}{c}{DINO}				{\includegraphics[height=2cm, width=2cm]{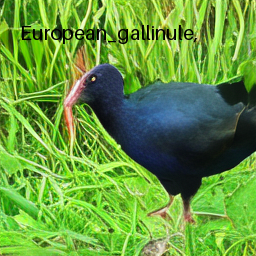}}&

					{\includegraphics[height=2cm, width=2cm]{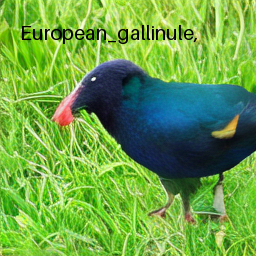}}&
					{\includegraphics[height=2cm, width=2cm]{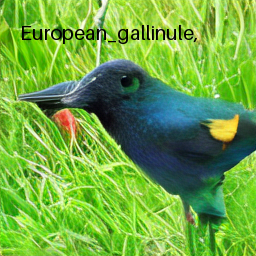}}\\
					\multicolumn{1}{c}{nearest}& \multicolumn{1}{c}{middle}&  \multicolumn{1}{c}{furthest}
					
			\end{tabular}}
			
			\vspace{-1em}
			\caption{Qualitative results showing the clean samples using clustering-based pruning for PR=0.5.}
			\vspace{-1.5em}
			\label{fig:clip_vs_dino}
		\end{figure}

	\myparagraph{Qualitative evaluation.}
	In figure \ref{fig:qual_imagenet} we visualise results for different pruning strategies and PRs. In accordance with the metric evaluation, we can distinguish that Cluster$_{C/D}$ achieves the best perceived quality which is also in accordance with the quantitative metrics that we show above. We include further qualitative results that showcase how the generated samples changes across the different methods for an identical noise vector in the supplementary. Like in CelebA-HQ, on ImageNet we also observe that the generated samples using different pruning strategies converge to similarly looking samples.  Figure~\ref{fig:clip_vs_dino} compares Cluster$_{C}$  versus Cluster$_{D}$ and  Cluster$_{C/D}^{mid}$   when PR=0.5. We see that the perceived quality is better in when using Cluster$_{D}$ and when the samples the samples selected are in closer proximity to the center.
	
		\subsection{Analysis}
\begin{figure}[]

			\hfill
			\parbox{\dimexpr\linewidth-1em}{%
				\resizebox{0.47\textwidth}{!}{%

					\begin{tabular}{cccc}

						{\includegraphics[height=2cm, width=2cm]{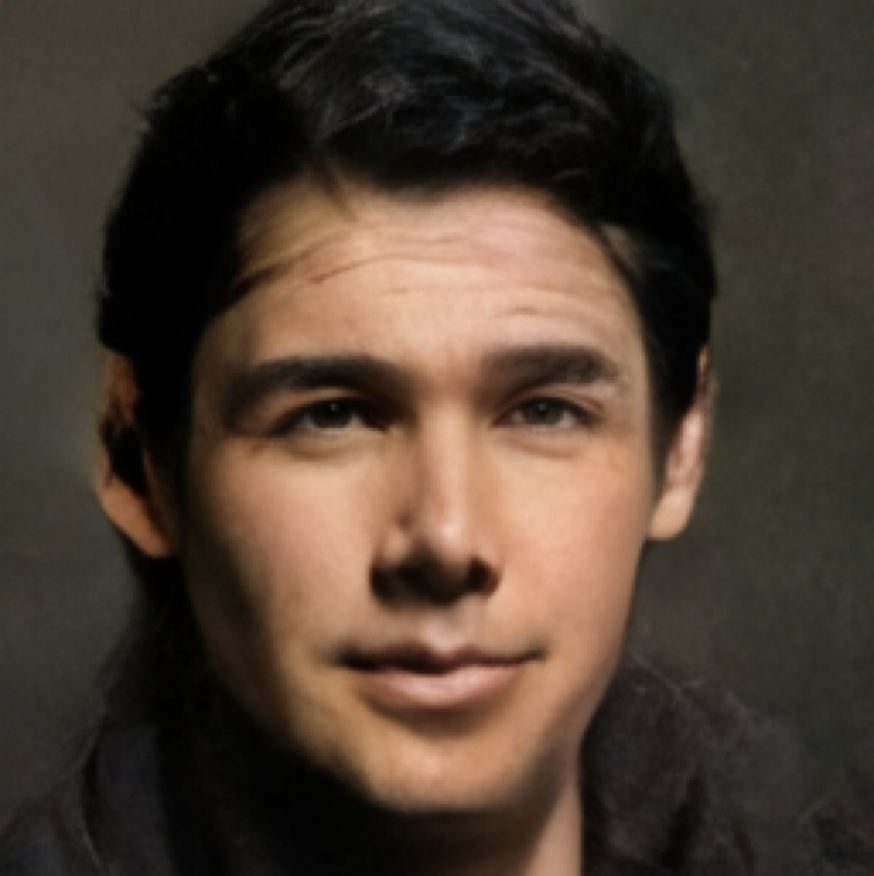}}&
						{\includegraphics[height=2cm, width=2cm]{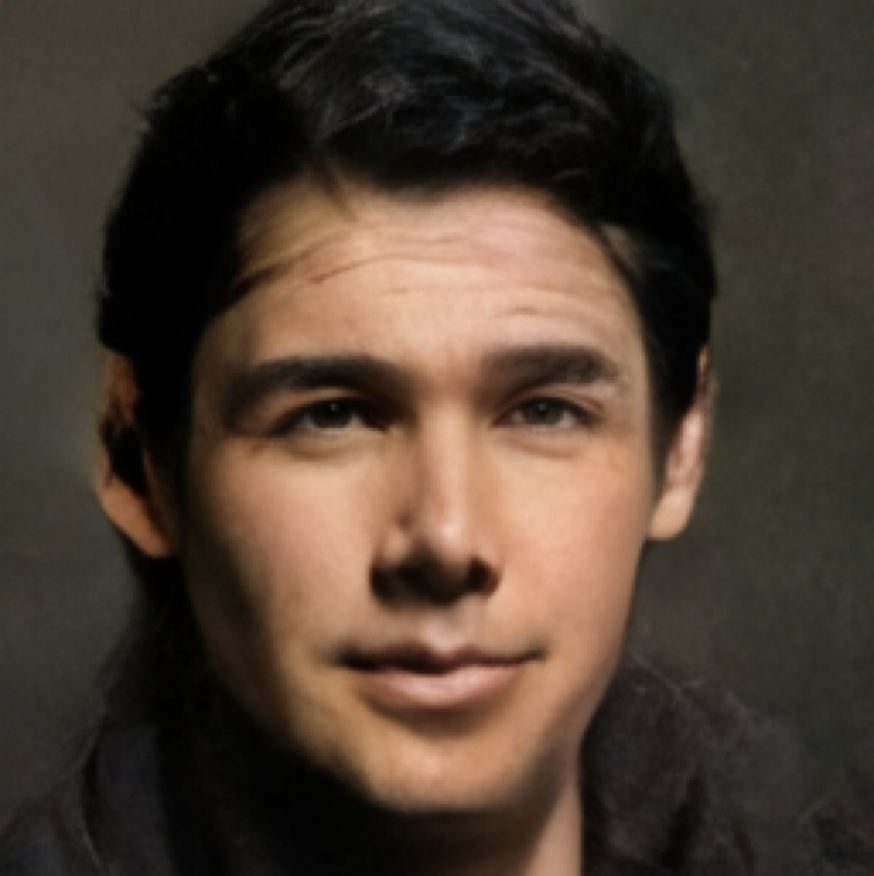}}&

						{\includegraphics[height=2cm, width=2cm]{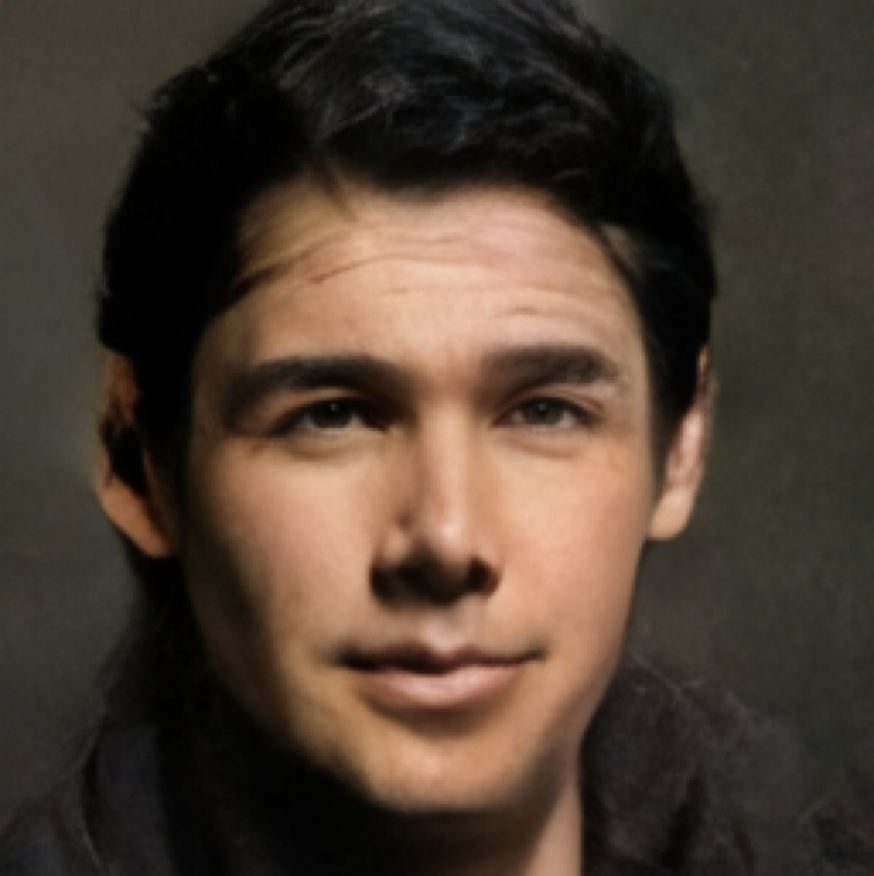}}&
						{\includegraphics[height=2cm, width=2cm]{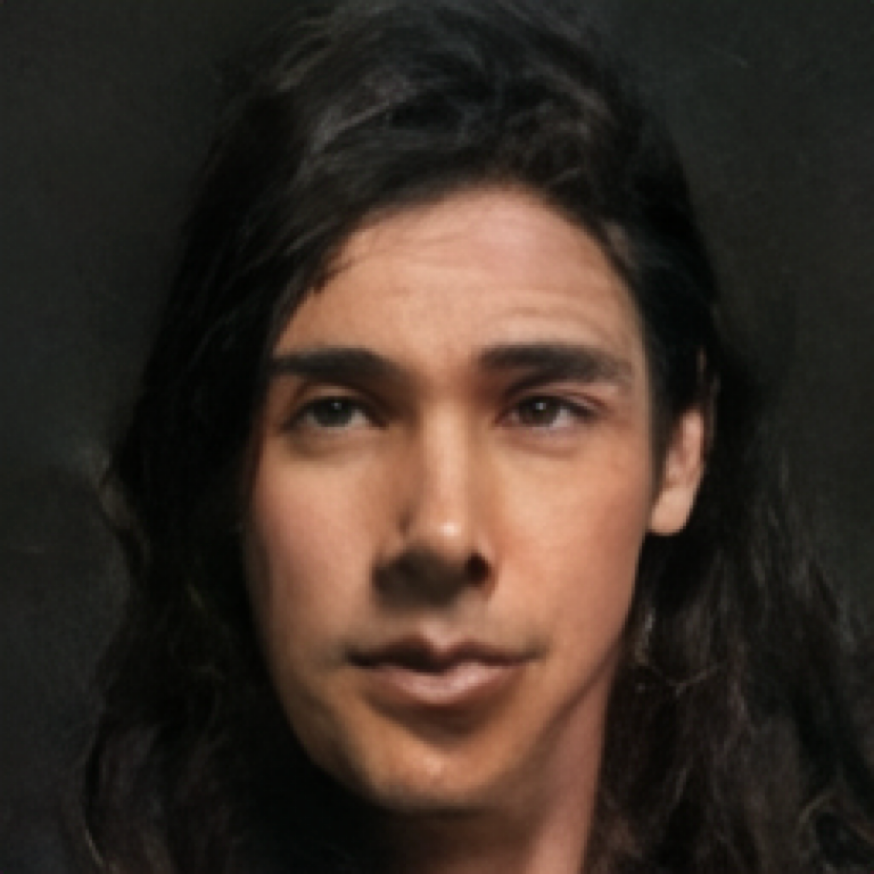}}
						\\
						\multicolumn{1}{c}{MoSo}& \multicolumn{1}{c}{MoSo$^{-1}$}&\multicolumn{1}{c}{GraNd}&	\multicolumn{1}{c}{GraNd$^{-1}$}
						
			\end{tabular}}}
			\vspace{-1em}
			\caption{Qualitative results showing the clean samples of MoSo and GraNd and their inverse}
			\vspace{-0.5em}
			\label{fig:qual_inv_disjoint}
		\end{figure}

		We observed some clear differences in the results on the two datasets which leads to the conclusion that the performance is linked to the distribution governing each dataset. For example, pruning based on clustering did not yield an advantage on CelebA-HQ and the random baseline nearly outperformed all other strategies. This is reasonable, a random baseline that selects samples uniformly at random covers the entire space of the distribution. On ImageNet, a much larger dataset, the positive effect of clustering was clearly visible, agreeing with our intution that samples in close proximity to the clusters centers are representative of the data distribution and are helpful in guiding the model learn the most dominant patterns. In contrast, when giving higher ranking to samples furthest away from the cluster, this had severly impacted the model. While these samples introduce more difficult or scarce samples, they are also inherently less representative of the dense region of the distribution, making it harder for the model to learn a coherent representation of the distribution. In this context, even the samples taken from the middle region of the cluster proved they were also not adequately represenative of the distribution. The fact that neither GraNd nor MoSo yielded better performance, despite being theoretically rationalized and despite their supply of a strong signal to the model during training, indicates that methods that do not involve density estimation are not particularly helpful to a generative model in learning the distribution. This is evident from the experiment over ImageNet, where clustering yielded the best performance.
		Some of the results that we obtained were counter-intuitive, for example, MoSo performing worse than GraNd despite it being more informed in its selection process.
		We also found it interesting that given the same initial noise vector, the model frequently (but not always) generates a similar image. In the case of human faces, we can regognize the identity of the face is the same despite slight differences in appearance. This happens even when the model is trained on two disjoint datasets, which happens when we train a model using a pruning strategy and its inverse with PR=0.5. In this case a method will retain 50\% of the highest-scoring sampless, while its inveres retains the remaining 50\% lowest-scoring samples. An example can be seen in figure~\ref{fig:qual_inv_disjoint}. We also see that frequently given the same initial noise vector, the final clean sample is similar, in this case however, there might be an overlap in the selected population between the different strategies. This shows that once a DMlearns to estimate a distribution, given a noise sample, the different models ultimately converge to samples that reside in the same reigion on the clean manifold. 
		
		\myparagraph{Comparison with other generative models.} We carried out an experiment in which we train a VAE and a GAN on a reduced dataset selected based on a random baseline, and the degradation in this case is instantateous. The resilience of DMs to data pruning in comparison is due to the iterative refinemenet using step-by-step denoising as the model gradually recovers the original data from noisy inputs, as opposed to GANs and VAEs which recover the data in a single pass thereby solving a much harder problem. Another pivotal factor in preserving the performance is the presence of the autoencoder, which allows the DM to operate in a lower more structured dimensional space. Having been trained on a more comprehesive dataset, the autoencoder can reinterpret the DM's output in a rich and representative latent space. 
		
\begin{figure}[]

			\hfill
			\parbox{\dimexpr\linewidth-1em}{%
				\resizebox{0.47\textwidth}{!}{%

					\begin{tabular}{cccc}

						{\includegraphics[height=2cm, width=2cm]{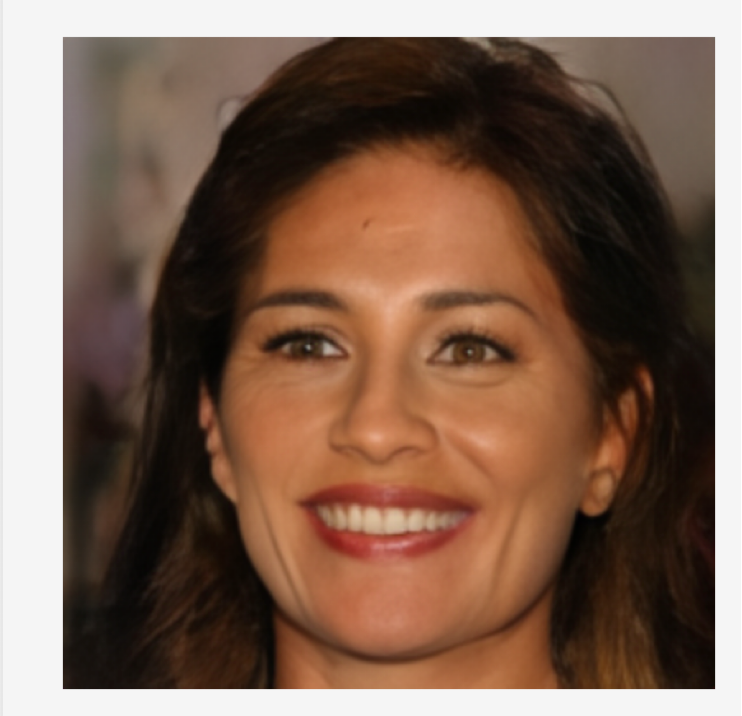}}&
						{\includegraphics[height=2cm, width=2cm]{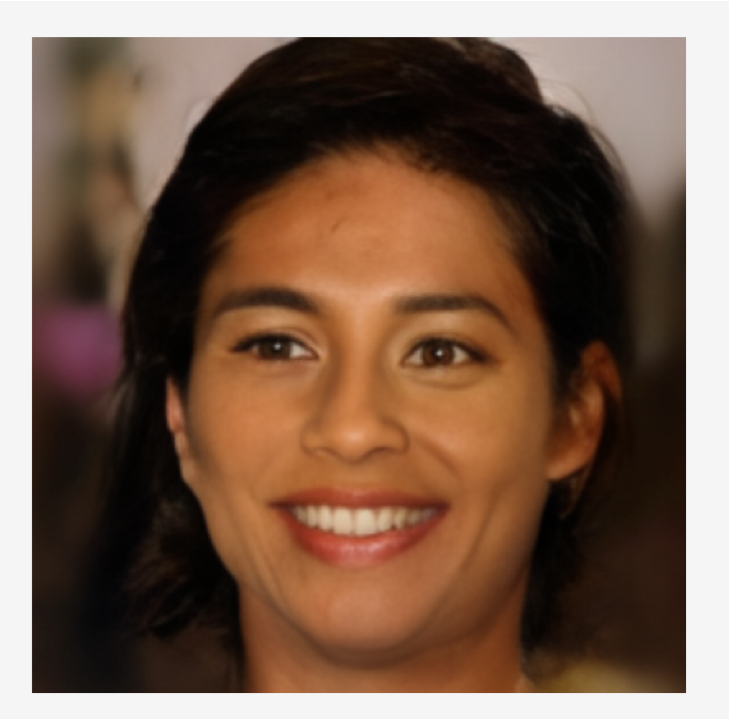}}&

						{\includegraphics[height=2cm, width=2cm]{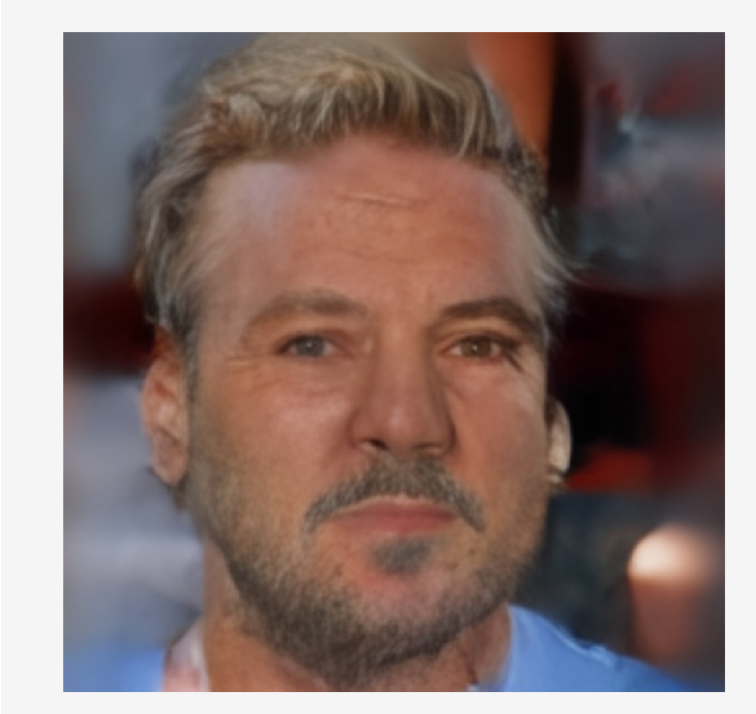}}&
						{\includegraphics[height=2cm, width=2cm]{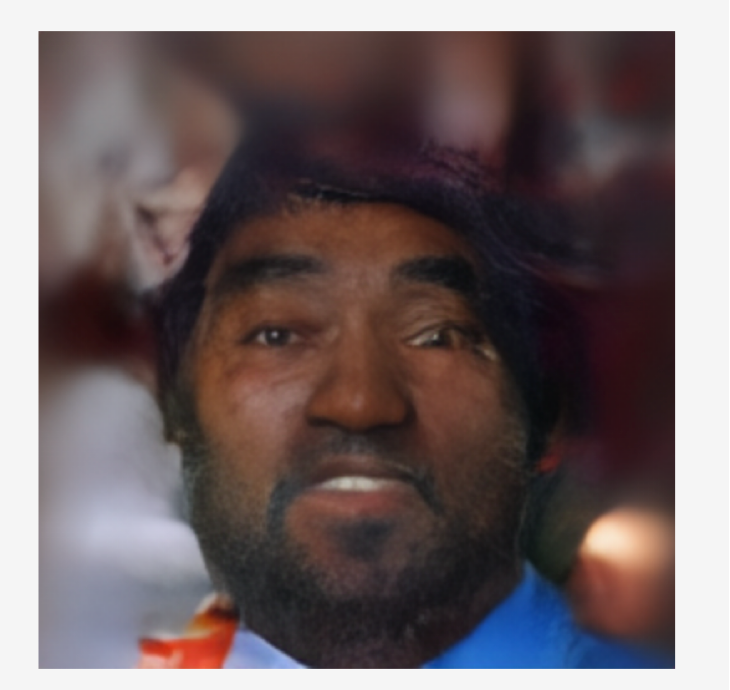}}
						\\
						\multicolumn{1}{c}{proportional}& \multicolumn{1}{c}{balanced }&\multicolumn{1}{c}{proportional}&  \multicolumn{1}{c}{balanced}
			\end{tabular}}}
			\vspace{-0.5em}
			\caption{Balanced sampling from clusters attempts to generate minority populations.}
			\vspace{-0.5em}
			\label{fig:proportional_balanced}
		\end{figure}

\begin{figure}
			\centering
			\captionsetup{justification=centering}
			\begin{subfigure}[]{0.45\textwidth}
				\includegraphics[width=1\linewidth]{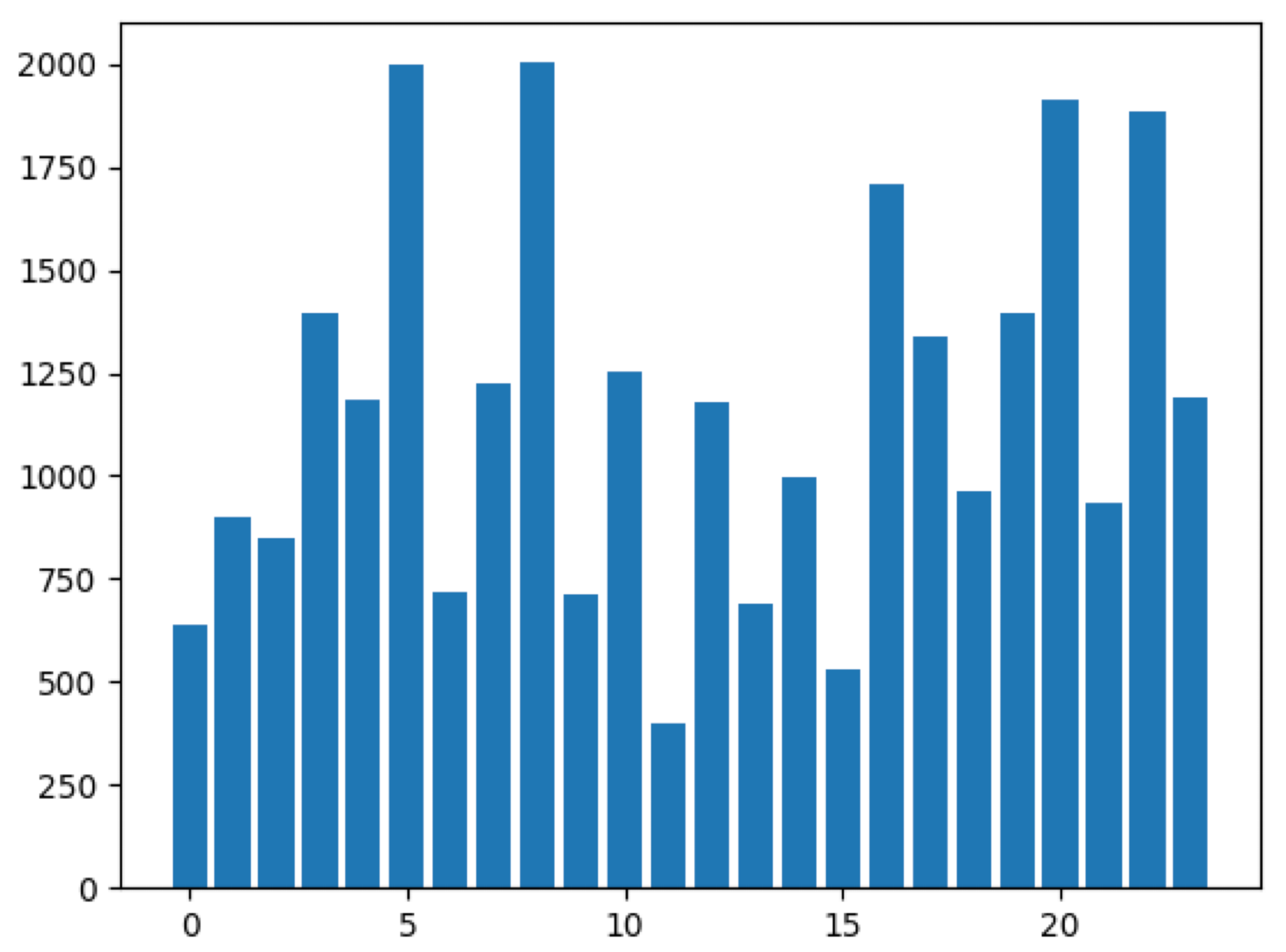}
				
				\caption{Clustering of CelebA-HQ training data.}
				
				\label{fig:bar_train_cluster}
			\end{subfigure}
			\begin{subfigure}[]{0.45\textwidth}
				\includegraphics[width=1\linewidth]{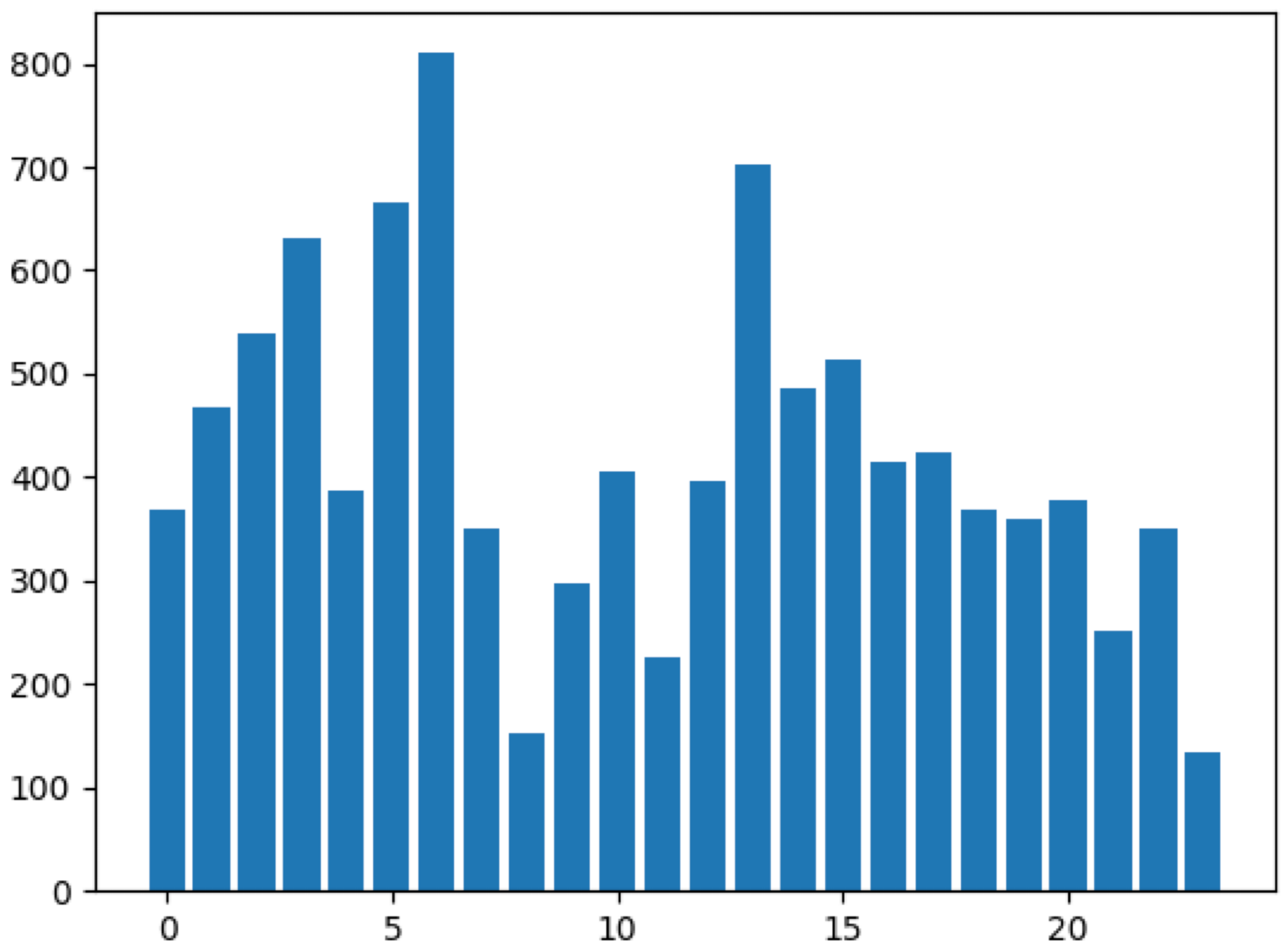}
				
				\caption{Clustering of the generated data when pruning proportionally to the cluster size.}
				
				\label{fig:bar_prop_gen_cluster}
			\end{subfigure}
			\begin{subfigure}[]{0.45\textwidth}
				\includegraphics[width=1\linewidth]{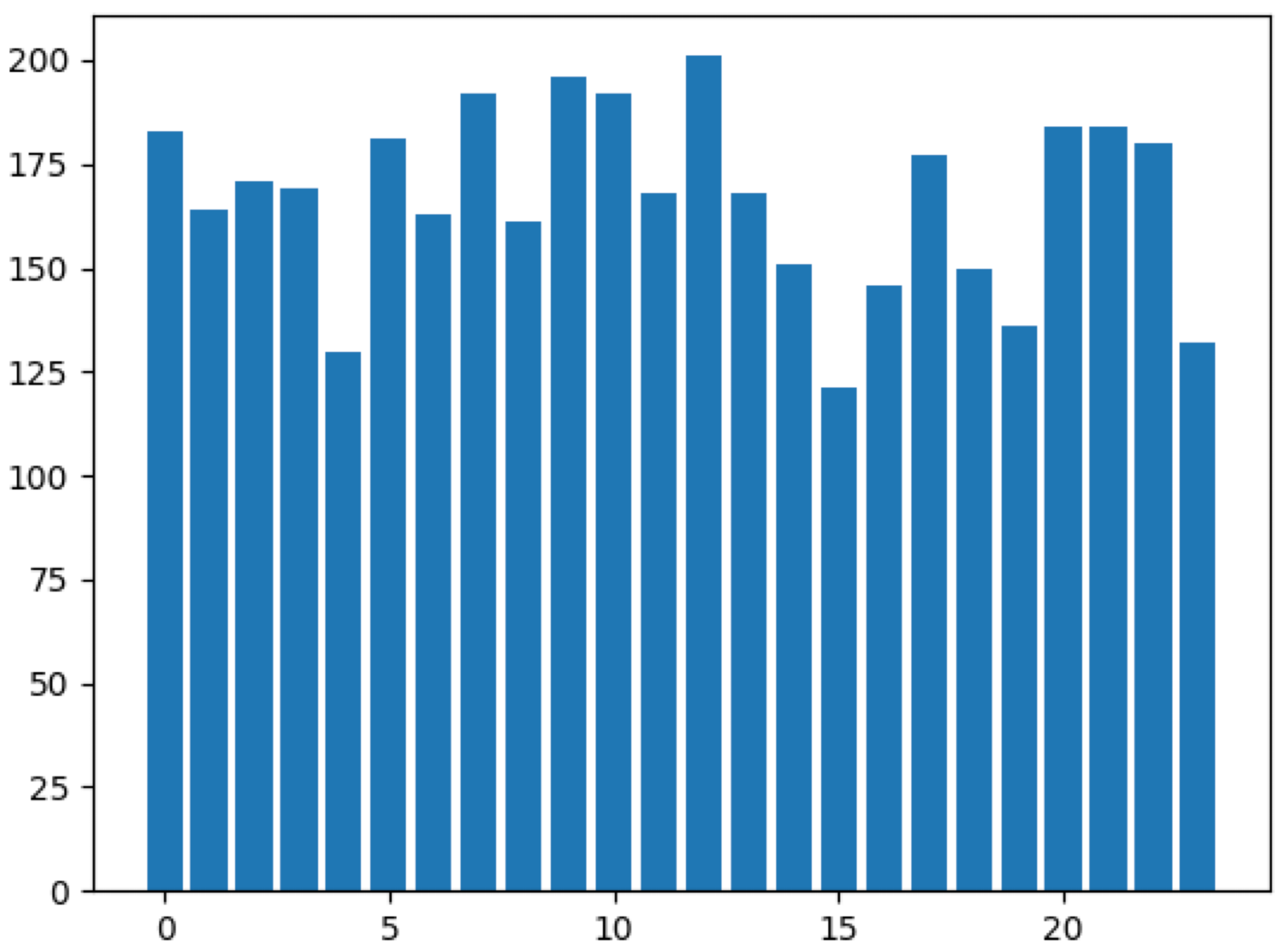}
				
				\caption{Clustering of CelebA-HQ generated data when pruning in a balanced manner.}
				
				\label{fig:bar_balanced_gen_cluster}
			\end{subfigure}
			
		\end{figure}

		\myparagraph{Balancing skewed distributions.} 
		In table~\ref{tab:equal_vs_proportioanl_clustering} we vary the proportion of samples selected from each cluster while keeping the total number of samples equal in both variants. In this experiment PR is determined by the number of samples in the smallest cluster, denoted by $s$, such that $PR=s\times k / |S| $ where $k$ is the number of clusters and S is the unpruned dataset. In \textit{proportional clustering} is the same as Cluster$_C$, in which we prune PR of the samples from each cluster, keeping the number of samples selected from each cluster proportional to its size. In \textit{Balanced clustering} we select an identical number of samples $s$ from each cluster. In CelebHQ this amounts to $\approx 70\%$  and  $\approx 85\%$ in CelebA-HQ and Imagenet correspondingly. In CelebA-HQ, although visually we observed that the generated samples are more balanced with respect to the underrepresented data, since we are computing the distance with respect to a skewed distribution it is reasonable that the FID becomes lower. In this case, the FID score is not a suitable measure. The balance can be observed in~\ref{fig:bar_balanced_gen_cluster} where the features of the generated samples forms more balanced groups as opposed to pruning based on proportional clustering~\ref{fig:bar_prop_gen_cluster}, which more aligns with the original distribution displayed in ~\ref{fig:bar_train_cluster}. In ImageNet, however, where the model was much more tolerant towards pruning especially with clustering-based pruning, the FID nearly remains constant.
		When training using balanced clusters, the model strives to increase the represetations of all samples, as can been seen in figure~\ref{fig:proportional_balanced} when compared to proportional clustering.

		\begin{table}[h!]
			\centering
			\begin{tabular}{lccc|c}
				\hline
				Method	& Proportional clustering&Balanced clustering \\
				\hline
				CelebA-HQ  &34.46 &42.63\\
				ImageNet & 12.82 & 12.88
				\vspace{-0.5em}
			\end{tabular}
			\vspace{-0.5em}
			\caption{Comparing FID when samples are selected proportionally to the cluster size and when balanced.} 
			\vspace{-1.5em}
			\label{tab:equal_vs_proportioanl_clustering}
		\end{table}
		
		\section{Conclusion}
		In this work we introducted data pruning to generative diffusion models. We have conducted a thorough experimental evaluation of the impact different pruning strategies have on the model. 
		Through a comprehensive experimental evaluation, we showed that data pruning is beneficial for training diffusion mdeosl. Our findings highlight the tolerant nature of DMs to data pruning, demonstrating that unlike other generative models, DMs can maintain high-quality outputs even with significantly reduced training datasets. This makes it possible to utilize DMs even when data is scarce. This study also emphasizes the potential for optimizing training efficiency and resource utilization in generative DMs without compromising their performance.
		
		\section*{Acknowledgement}
		This work is funded by the German Federal Ministry for Economic Affairs and Climate Action within the project ”nxtAIM”. This work was futher supported by Helmholtz AI computing resources (HAICORE) of the Helmholtz Association’s Initiative and Networking Fund through Helmholtz AI. Computational resources were also provided by the German AI Service Center WestAI.

		%-------------------------------------------------------------------------
		
		%-------------------------------------------------------------------------
		\clearpage
		%%%%%%%%% REFERENCES
		{\small
			\bibliographystyle{ieee_fullname}
			\bibliography{egbib}
		}
		
	\end{document}